\documentclass{article}

\PassOptionsToPackage{table}{xcolor}
\usepackage[nonatbib,preprint]{neurips_2026}
\usepackage[numbers,sort&compress]{natbib}

\usepackage[utf8]{inputenc}
\usepackage[T1]{fontenc}
\usepackage{hyperref}
\usepackage{url}
\usepackage{booktabs}
\usepackage{amsfonts}
\usepackage{microtype}
\usepackage{xcolor}
\usepackage{graphicx}
\usepackage{wrapfig}
\usepackage{capt-of}
\usepackage{amsmath}
\usepackage{multirow}
\usepackage{longtable}
\usepackage{array}
\usepackage{pifont}
\usepackage{fontawesome5}
\usepackage{inconsolata}
\usepackage[most]{tcolorbox}
\usepackage{enumitem}
\usepackage{fvextra}

\makeatletter
\renewcommand{\@notice}{}
\makeatother

\newcommand{\benchname}{\textsc{SkillFlow}}
\definecolor{bettermore}{RGB}{226,239,218}
\definecolor{betterless}{RGB}{221,235,247}
\newcommand{\cmark}{\textcolor{green!60!black}{\ding{51}}}
\newcommand{\xmark}{\textcolor{red!70!black}{\ding{55}}}

\definecolor{usercolor}{RGB}{70, 130, 180}
\definecolor{assistantcolor}{RGB}{34, 139, 34}
\definecolor{toolcolor}{RGB}{255, 140, 0}
\definecolor{systemcolor}{RGB}{128, 128, 128}
\definecolor{instructioncolor}{RGB}{138, 43, 226}

\definecolor{diffred}{RGB}{180,40,40}
\definecolor{diffgreen}{RGB}{30,140,60}
\definecolor{diffmeta}{RGB}{90,90,90}
\definecolor{diffbg}{RGB}{248,248,248}

\newtcolorbox{userbox}{
    colback=blue!5,
    colframe=usercolor,
    boxrule=1pt,
    arc=3pt,
    left=5pt,
    right=5pt,
    top=3pt,
    bottom=3pt,
    fonttitle=\bfseries,
    title=User,
    breakable,
    before skip=2pt,
    after skip=2pt
}

\newtcolorbox{assistantbox}{
    colback=green!5,
    colframe=assistantcolor,
    boxrule=1pt,
    arc=3pt,
    left=5pt,
    right=5pt,
    top=3pt,
    bottom=3pt,
    fonttitle=\bfseries,
    title=Assistant,
    breakable,
    before skip=2pt,
    after skip=2pt
}

\newtcolorbox{toolbox}{
    colback=orange!5,
    colframe=toolcolor,
    boxrule=1pt,
    arc=3pt,
    left=5pt,
    right=5pt,
    top=3pt,
    bottom=3pt,
    fonttitle=\bfseries,
    title=Tool Call/Response,
    breakable,
    before skip=2pt,
    after skip=2pt
}

\newtcolorbox{systembox}{
    colback=gray!5,
    colframe=systemcolor,
    boxrule=1pt,
    arc=3pt,
    left=5pt,
    right=5pt,
    top=3pt,
    bottom=3pt,
    fonttitle=\bfseries,
    title=System,
    breakable,
    before skip=2pt,
    after skip=2pt
}

\newtcolorbox{instructionbox}{
    colback=purple!5,
    colframe=instructioncolor,
    boxrule=1pt,
    arc=3pt,
    left=5pt,
    right=5pt,
    top=3pt,
    bottom=3pt,
    fonttitle=\bfseries,
    title=Instruction,
    breakable,
    before skip=2pt,
    after skip=2pt
}

\title{\raisebox{-0.2\height}{\includegraphics[height=1.6em]{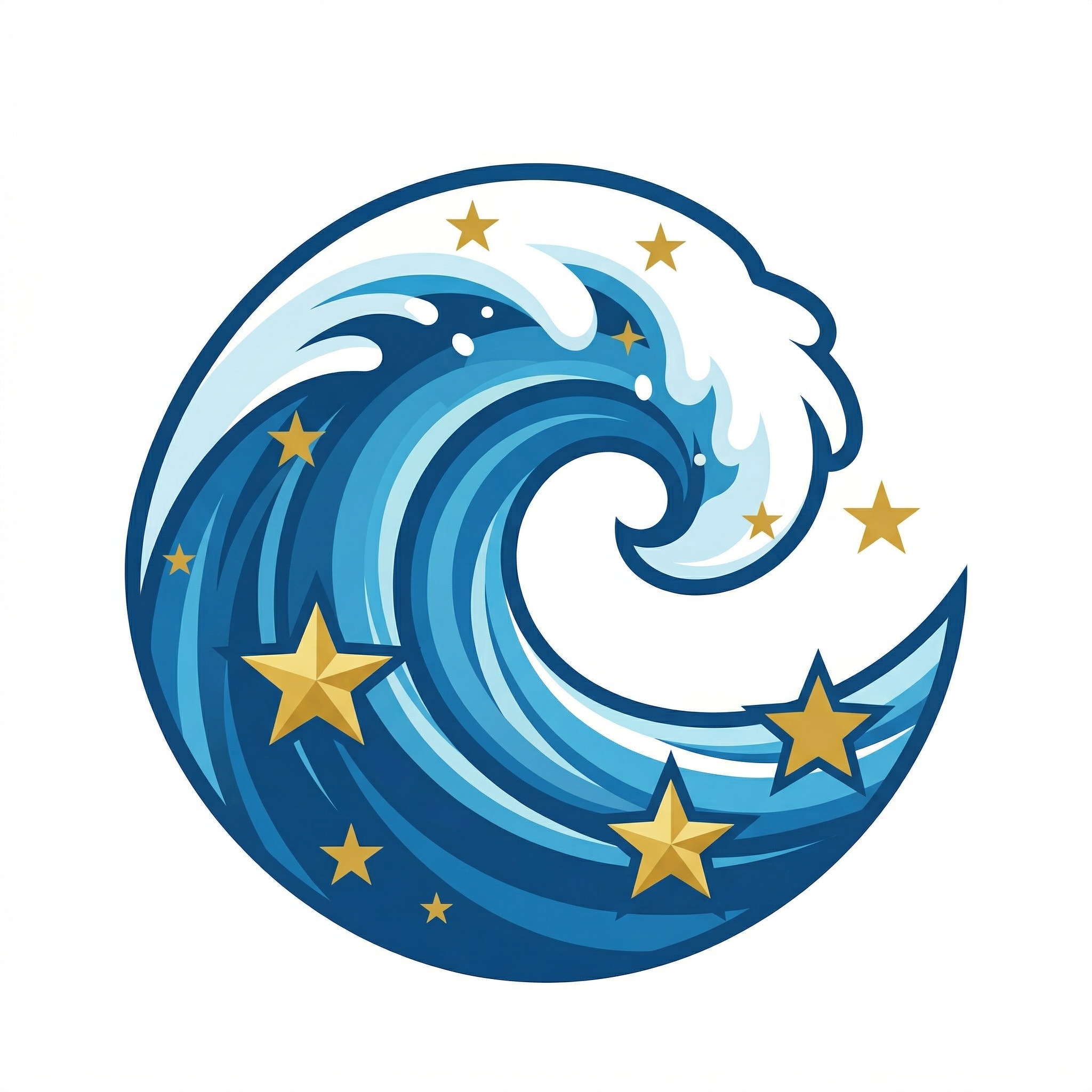}}\hspace{0.4em}\benchname: Benchmarking Lifelong Skill Discovery and Evolution for Autonomous Agents}

\author{Ziao Zhang$^{1}$ \quad Kou Shi$^{1}$ \quad Shiting Huang$^{1}$ \quad Avery Nie$^{2}$ \quad Yu Zeng$^{1}$ \quad Yiming Zhao$^{1}$ \\
\textbf{Zhen Fang$^{1}$ \quad Qisheng Su$^{1}$ \quad Haibo Qiu$^{3}$ \quad Wei Yang$^{1}$ \quad Qingnan Ren$^{1}$ \quad Shun Zou$^{1}$} \\
\textbf{\quad Wenxuan Huang$^{1}$ \quad Lin Chen$^{1}$ \quad Zehui Chen$^{1}$ \quad  Feng Zhao$^{1}$\thanks{\quad Corresponding author}}\\
$^{1}$University of Science and Technology of China \\
$^{2}$University of Toronto \quad $^{3}$University of Sydney
\\
\vspace*{0.4cm}
\centering
\urlstyle{same}
\href{https://zhangzi-a.github.io/SkillFlow-project-page/}{\faGlobe\ \textbf{Project Page}}
\vspace{0.15cm}}

\begin{document}

\maketitle

\begin{abstract}

As the capability frontier of autonomous agents continues to expand, they are increasingly able to complete specialized tasks through plug-and-play external skills. Yet current benchmarks mostly test whether models can use provided skills, leaving open whether they can discover skills from experience, repair them after failure, and maintain a coherent library over time. We introduce \textbf{\benchname{}}, a benchmark of 166 tasks across 20 families in which task construction within each family follows a Domain-Agnostic Execution Flow (DAEF) that defines an agent workflow framework, allowing these tasks to share a consistent workflow. Agents are evaluated under an Agentic Lifelong Learning protocol in which they begin without skills, solve tasks sequentially within each family, externalize lessons through trajectory- and rubric-driven skill patches, and carry the updated library forward. Experiments reveal a substantial capability gap. For Claude Opus 4.6, lifelong skill evolution improves task success from 62.65\% to 71.08\% (+8.43 points). However, high skill usage does not necessarily imply high utility: Kimi K2.5 gains only +0.60 points despite 66.87\% skill usage, while Qwen-Coder-Next reaches only a 44.58\% task completion rate and still regresses relative to the vanilla setting. \benchname{} contributes a structured testbed for this direction and an in-depth empirical analysis of skill discovery, patching, transfer, and their failure modes under lifelong evaluation. The code is available at
\href{https://github.com/ZhangZi-a/SkillFlow}{https://github.com/ZhangZi-a/SkillFlow}.

\end{abstract}

\begin{figure*}[t]
    \centering
    \includegraphics[width=\textwidth]{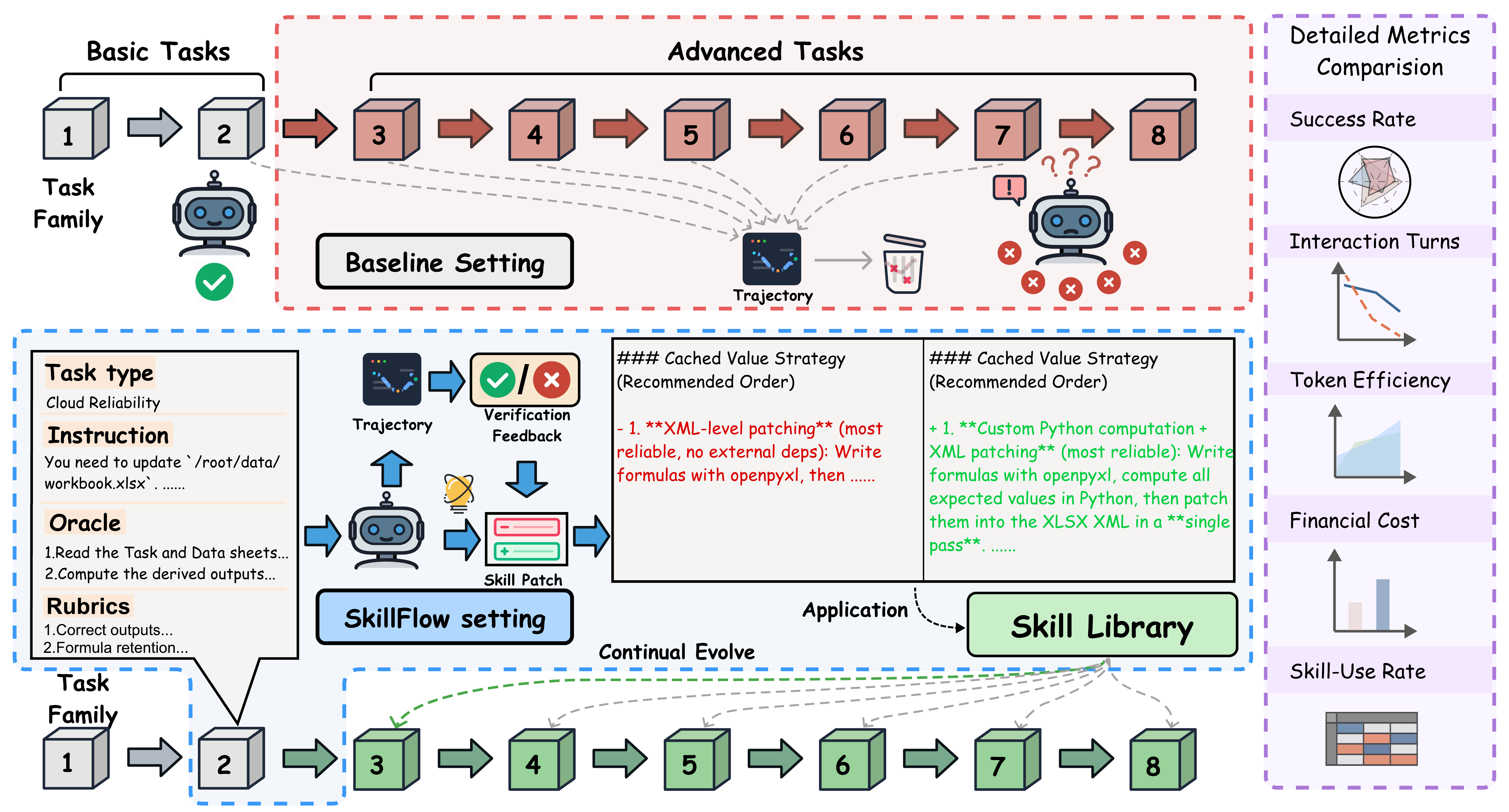}
    \caption{Conceptual Overview of \benchname. The figure contrasts conventional static-skill evaluation with our lifelong setting, in which agents externalize experience into reusable skill artifacts, revise them through patches, and transfer them across tasks that share a common DAEF.}
    \label{fig:teaser}
\end{figure*}

\begin{figure*}[t]
    \centering
    \includegraphics[width=\textwidth]{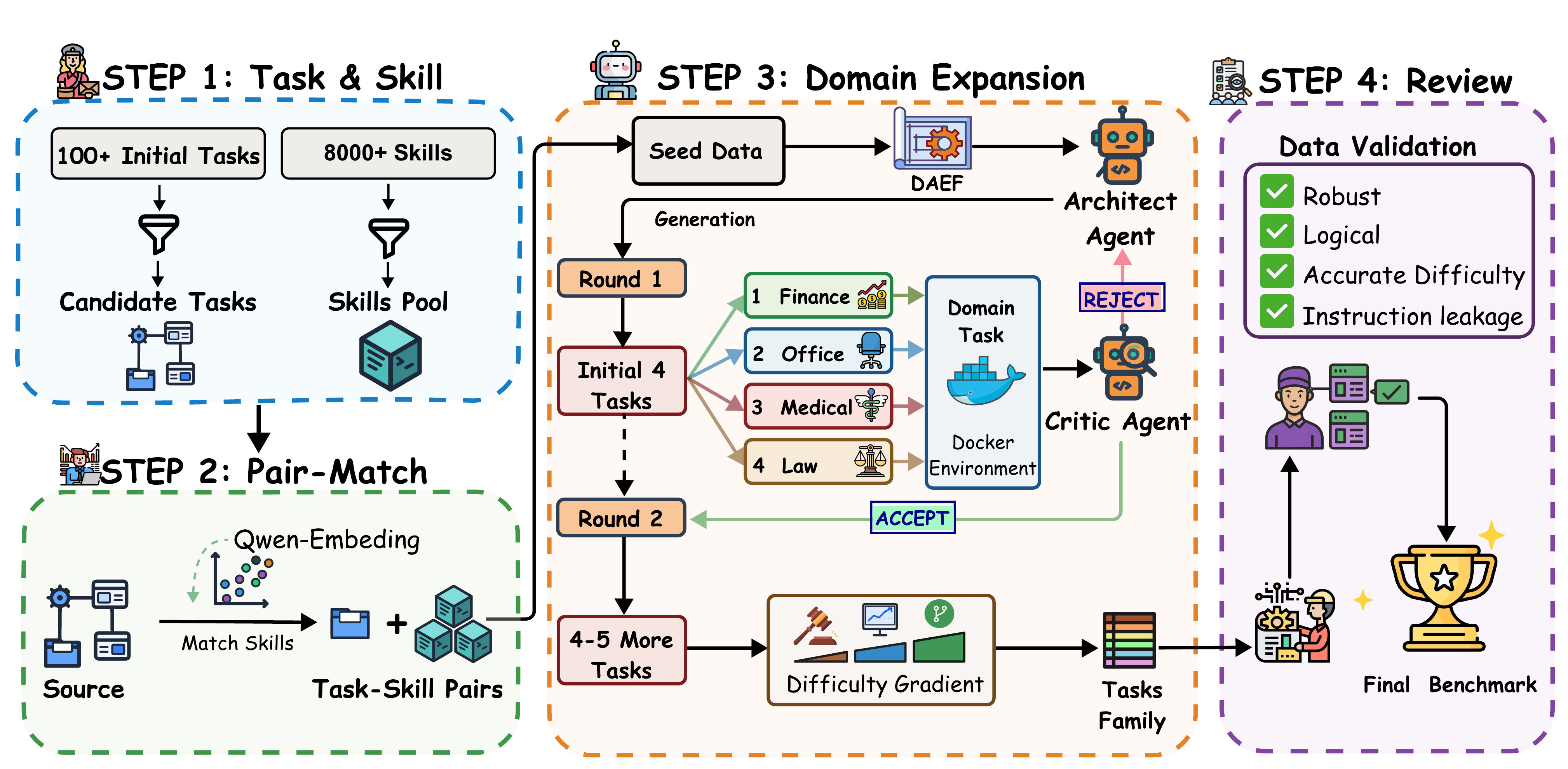}
    \caption{Task-Construction Pipeline of \benchname. Step 1 collects candidate seed tasks and a curated external skill pool. Step 2 uses embedding-based pair matching to attach relevant skills to each seed and form task--skill reference pairs. Step 3 performs domain expansion under a fixed DAEF: the Architect Agent generates an initial cross-domain task set, the Critic Agent either rejects or accepts the family, and accepted families proceed to a second expansion round that adds more tasks and establishes a difficulty gradient. Step 4 applies human review over robustness, logical soundness, difficulty calibration, and instruction-leakage risk before final benchmark inclusion.}
    \label{fig:pipeline}
\end{figure*}

\section{Introduction}

Frontier Large Language Models(LLMs) systems are increasingly deployed as autonomous command-line agents that solve realistic multi-step tasks in terminal environments. Claude Code~\cite{anthropic2025claudecode}, Gemini CLI~\cite{google2025geminicli}, Codex CLI~\cite{openai2025codexcli}, and Qwen-Coder~\cite{qwen_code} are prominent examples. Agent skills have recently emerged as external packages of procedural knowledge that augment model capabilities for specialized tasks. These skills can encode usage scenarios, core guidance, and reusable code or documentation assets. As a result, many vendors are beginning to release agent versions with native skill support.

SkillsBench~\cite{li2026skillsbench} provides an early demonstration that skills can substantially assist models in solving tasks, while also suggesting that current models remain limited in generating skills autonomously. However, it does not examine whether an LLM can discover and summarize reusable skills from its own task-solving process and apply them to future tasks of the same type. SkillWeaver~\cite{zheng2025skillweaver}, SkillRL~\cite{xia2026skillrl}, MemSkill~\cite{zhang2026memskill}, and related systems~\cite{zhou2026memento} also show that experience-derived skills can improve downstream performance. This leads to a central question: \textbf{can an autonomous agent extract reusable skills from its own experience, repair them after failures, and maintain an evolving skill library across a sequence of tasks?} To answer this question, we introduce \textbf{\benchname}, a benchmark of 166 runnable tasks across 20 task families for measuring lifelong skill discovery and evolution.

To construct the benchmark, we first analyze existing agent loops and find that mature agents~\cite{huang2026vision,zeng2026vision, fang2025dualvla} typically rely on stable AI workflows, which allow them to solve tasks of the same type in a consistent manner. Based on this observation, we define the Domain-Agnostic Execution Flow (DAEF), a workflow framework that preserves operational topology while abstracting away domain-specific entities and serves as a scaffold for benchmark construction. We use this structure to support controlled progression within each task family. Concretely, we select realistic tasks from GDPval~\cite{patwardhan2025gdpval} and SkillsBench~\cite{li2026skillsbench} and extract a DAEF from each seed task. We then instantiate each DAEF through a dual-agent iterative pipeline using the original task as a reference instance, where one agent constructs task assets and the other reviews the generated tasks and provides timely feedback for revision. This process yields 8 to 9 tasks in the Harbor format~\cite{Harbor_Framework} for each DAEF, which together form a task family. After human verification, the final benchmark contains 20 task families and 166 tasks. Evaluation then follows an \textbf{Agentic Lifelong Learning} protocol in which an agent begins without skills, solves tasks sequentially within each family, summarizes lessons from trajectories and rubric feedback, and updates the library through explicit \textit{skill patches} after each task.

Experiments reveal a substantial \textbf{capability gap}. On Claude Opus 4.6, lifelong skill evolution improves task success from 62.65\% to 71.08\% (+8.43 points). Yet high usage does not imply high utility: Kimi K2.5 gains only +0.60 points despite a 66.87\% skill usage rate, while Qwen-Coder-Next reaches only a 44.58\% task completion rate and still regresses relative to the vanilla setting. We find that most current models fail to achieve stable self-evolution through iterative skill updates. Stronger models tend to consolidate prior experience into stable workflows and continue refining them over time, thereby realizing sustained capability gains. Weaker models struggle to integrate multiple experiences into a coherent procedure, which leads to cognitive overload in subsequent tasks.

\paragraph{Contributions.}
Our contributions are listed:
\begin{itemize}
    \item We introduce \textbf{\benchname{}}, a benchmark of 166 runnable tasks across 20 task families for evaluating lifelong skill discovery and evolution in autonomous agents.
    \item We design \textbf{task families with a shared agent workflow framework} based on DAEF, enabling systematic evaluation of cross-task skill transfer learning.
    \item We specify an \textbf{Agentic Lifelong Learning} protocol in which agents begin without skills, evolve an external skill library through trajectory- and rubric-driven \textit{skill patches}, and progressively learn and are evaluated under a fixed task order within each family.
    \item We identify and closely analyze \textbf{common failure patterns} in skill evolution, including fragmented skill growth, reinforcement of erroneous logic, and the gap between writing skills and reliably repairing them, revealing current models' limitations in producing compact, transferable procedural skills.
\end{itemize}

\section{\benchname{}}

\subsection{Benchmark Overview}

\benchname{} contains 20 workflow families and 166 tasks spanning five broad domains: \textit{Finance \& Economics}, \textit{Operations \& Supply Chain}, \textit{Healthcare \& Life Sciences}, \textit{Governance \& Strategy}, and \textit{Data \& Document Intelligence}. Each benchmark instance is defined by a family-local task sequence, a fixed within-family difficulty order, and an associated verifier style. Family sizes range from 8 to 9 tasks. Overall, the benchmark encompasses a range of realistic workplace workflows, including spreadsheet-centric planning, OCR and PDF extraction, office-document editing, compliance analysis, and quantitative reasoning.

\subsection{Domain-Agnostic Execution Flow}

The central design principle of \benchname{} is a \textbf{Domain-Agnostic Execution Flow (DAEF)}: a workflow skeleton shared by a family of tasks after removing domain-specific entities, file names, and business semantics. Formally, we represent a task instance as a domain-grounded workflow graph
\begin{equation}
\mathcal{T} = (V, E, \lambda, \gamma),
\end{equation}
where $V$ is a set of executable sub-goals or operations, $E \subseteq V \times V$ is a set of precedence or dependency edges, $\lambda(v)$ assigns each node a domain-agnostic operation type, and $\gamma(v)$ provides task-specific grounding, such as concrete files, entities, fields, or business objects.

A DAEF is the abstract workflow graph obtained by removing task-specific grounding while preserving operation types and dependency structure:
\begin{equation}
\mathcal{F} = \phi(\mathcal{T}) = (V_F, E_F, \lambda_F).
\end{equation}

\begin{figure*}[t]
    \centering
    \begin{minipage}[t]{0.5\textwidth}
        \vspace{0pt}
        \centering
        \vspace{6mm}
        \includegraphics[width=0.9\linewidth,height=0.24\textheight,keepaspectratio]{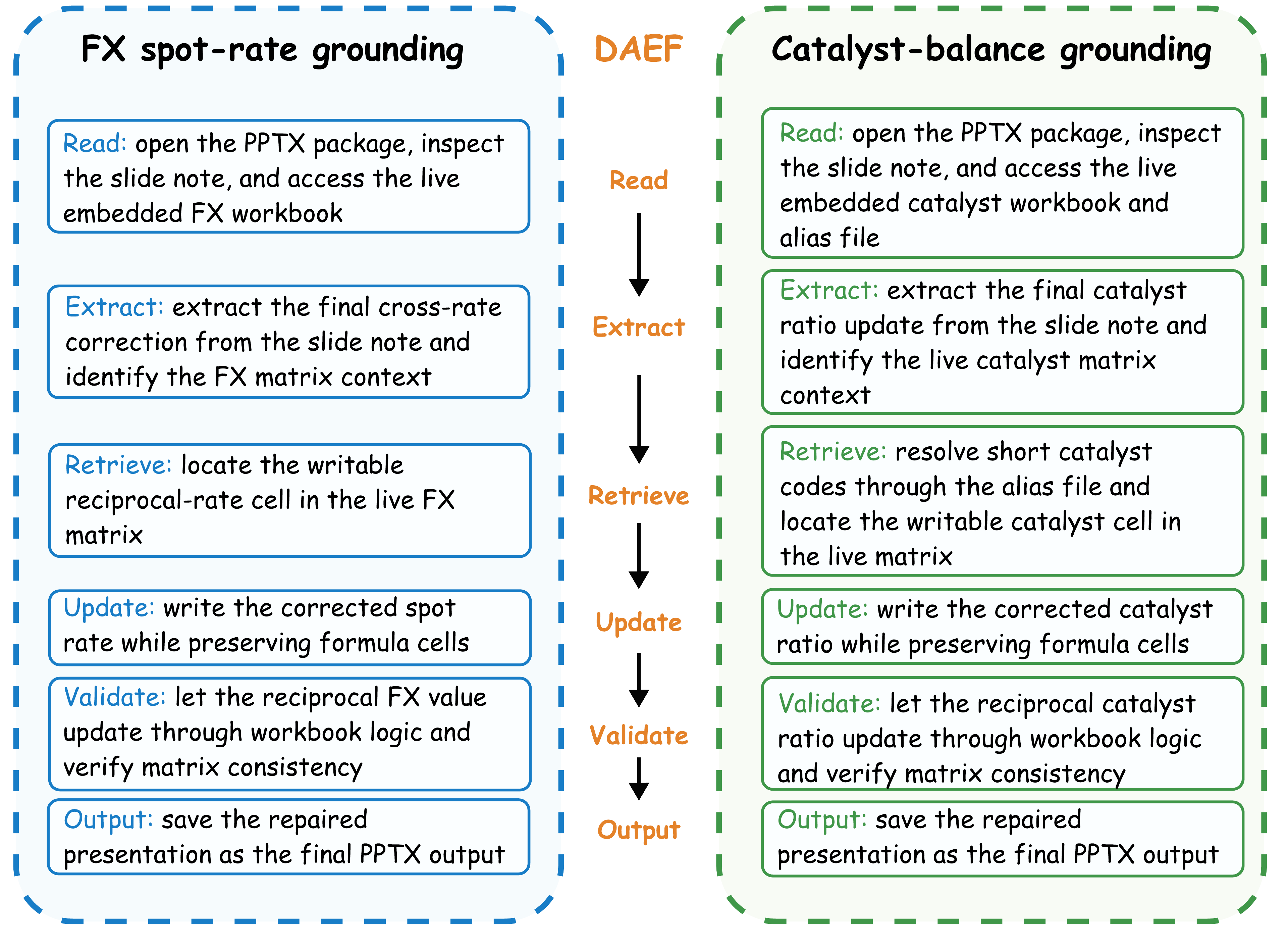}
        \captionof{figure}{DAEF correspondence across domains. Distinct tasks can instantiate the same abstract workflow, enabling cross-domain skill transfer.}
        \label{fig:daef-example}
    \end{minipage}\hfill
    \begin{minipage}[t]{0.48\textwidth}
        \vspace{0pt}
        \centering
        \includegraphics[width=0.9\linewidth,height=0.24\textheight,keepaspectratio]{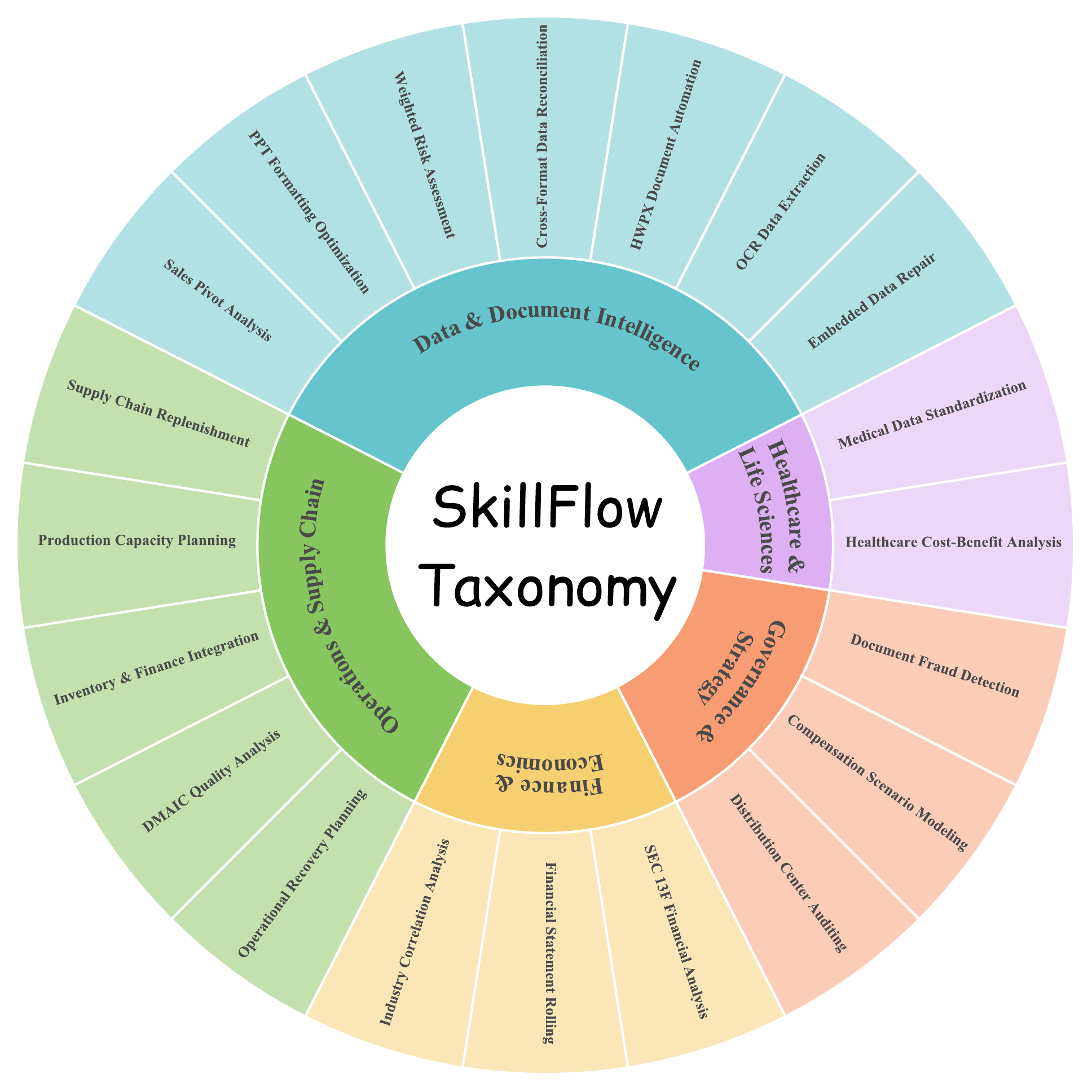}
        \captionof{figure}{Relationship Between Benchmark Categories and DAEFs. The inner ring shows the five top-level benchmark categories, and the outer ring shows the corresponding DAEF names under each category.}
        \label{fig:sunburst}
    \end{minipage}
\end{figure*}

\paragraph{Operationalizing DAEF in construction.}
We operationalize DAEF through a two-stage human annotation and standardization protocol that makes extraction explicit, standardized, and verifiable before task generalization.

\textbf{Stage 1: meta-step extraction.} Each seed task is independently annotated with a sequence of meta-step nodes containing 5--8 executable operations, dependency edges, and a short textual rationale. Node types are drawn from a controlled single-word vocabulary such as \textit{read}, \textit{retrieve}, \textit{compute}, \textit{detect}, and \textit{output}; the full inventory is listed in Appendix~\ref{sec:appendix-daef-vocab}.

\textbf{Stage 2: workflow construction.} The extracted workflow representation is normalized by collapsing task-specific grounding, standardizing node labels, and retaining only the dependency structure and operation inventory stable across tasks. The resulting canonical DAEF also specifies the allowed categories of variation to control the difficulty gradient of subsequently instantiated tasks. We provide the detailed annotation rules and expert agreement screening procedure in Appendix~\ref{sec:appendix-daef-vocab}. A canonical DAEF is retained only if it remains internally coherent and supports controlled generalization to new tasks without introducing a new workflow; a detailed table of allowed variation types appears in Appendix~\ref{sec:appendix-daef-variation}.

\subsection{Task Construction Protocol}

Figure~\ref{fig:pipeline} summarizes the process used to build \benchname{}. We organize it into \textbf{four steps}.

\paragraph{Step 1: seed task collection and skill curation.}
We identify seed tasks from two benchmark suites, retaining high-quality examples that reflect realistic workplace scenarios without requiring external network services. We exclude tasks that are overly narrow, simple, or difficult to verify reliably. This yields 64 seed tasks: 18 from SkillsBench~\cite{li2026skillsbench} and 46 from GDPval~\cite{patwardhan2025gdpval}. Simultaneously, we gather over 8,000 open-source skills from public repositories~\cite{claude_skills,openclaw_skills,anthropic_skills,skillsmap}, filtering them for safety, formatting, and domain relevance to obtain 2,318 skills for task construction.

\paragraph{Step 2: task--skill pair matching.}
To provide specialized support for subsequent task construction, we match each seed task with reference skills. Using Qwen3-embedding-4B~\cite{zhang2025qwen3}, we calculate semantic similarity between task and skill descriptions and retrieve 5--10 candidate skills for each seed, yielding 64 matched task--skill pairs.

\paragraph{Step 3: iterative task-family generalization.}
We then construct the family-level reference package used for domain expansion. For each selected seed, human annotators convert the task into a DAEF by extracting meta-steps and defining the workflow according to the previously described protocol. After deduplication and a generalization-difficulty check, we retain 30 DAEFs. The resulting seed-task reference package---including the original assets, matched skills, files, environment, and stable DAEF---enters an automated construction loop. We discard seeds at this stage if the resulting family fails to achieve execution stability or DAEF consistency within five revision iterations.

\textbf{(a) Architect Agent generation.}
The first agent, built within Cursor~\cite{cursor} using GPT-5.3-Codex~\cite{gpt53_codex}, serves as the Architect Agent for the new task family. Conditioned on the target DAEF, it proposes new task types and uses the seed task as a reference to construct task goals, create task assets, and define verification schemes for these tasks. With support from the matched skills, it then builds the corresponding Docker task environments and develops solver and verifier assets, producing four initial tasks in the Harbor format.

\textbf{(b) Critic Agent review.}
A second agent, employing Claude Opus 4.6, serves as the Critic Agent and evaluates the four initial tasks. It verifies and evaluates these tasks in real Docker environments and provides detailed feedback after jointly assessing workflow consistency, the presence of a difficulty gradient, task solvability, verifier correctness, and environment reliability.

\textbf{(c) Revision loop and expansion phase.}
The Architect Agent and the Critic Agent interact for multiple rounds until the Critic Agent approves the family. If the Critic Agent rejects the family more than five times, construction of that task family is abandoned. After passing inspection, the process enters a second round of task construction, which follows the same construction--verification loop as the first round and generates 4--5 new tasks. This two-round process produces a task sequence of 8--9 tasks for each family and ultimately yields 20 task families.

\paragraph{Step 4: human review and family revision.}
Human reviewers inspect each candidate family along four dimensions: instruction leakage, logical soundness of the task, correctness of the final task environment, and reasonableness of the difficulty gradient. Families with issues in any dimension are returned for manual revision. In practice, thanks to the strict construction procedure in the preceding steps, none of the 20 task families exhibits fatal environment errors, and all 20 families are retained after manual revision.

After the final review and revision process, this pipeline yields 20 workflow families and 166 tasks.

\subsection{Agentic Lifelong Learning Protocol}

We formalize the lifelong learning protocol as a sequential process over a family of tasks. Let $\mathcal{F} = \{T_1, T_2, \ldots, T_n\}$ denote an ordered task family, where tasks are sorted by within-family difficulty. The agent maintains an updatable skill library $\mathcal{S}_t$ at each step $t$. The protocol proceeds as follows:

\textbf{First-task execution and feedback.} For the first task $T_1$, the agent does not use any skill and completes the task under its native agent harness, producing an execution trace $\tau_1$. After task completion, the model receives a verifier-derived rubric $r_1$, which is a normalized textual description of missing or incorrect content.

\textbf{Skill patch generation.} After each task, the model learns from the resulting trajectory and verifier feedback and generates a skill patch conditioned on a fixed skill-patch prompt template $g$. This patch-generation step relies on the model's native capability rather than the surrounding agent harness. For the first task, the patch bootstraps the skill library from scratch:
\begin{equation}
\Delta_1 = \text{Model}_{g}(\emptyset, \tau_1, r_1), \quad \mathcal{S}_1 = \text{Apply}(\Delta_1, \emptyset).
\end{equation}
For all subsequent tasks $t > 1$, the agent executes $T_t$ with the current library $\mathcal{S}_{t-1}$, obtains an execution trace $\tau_t$ and verifier-derived rubric $r_t$, and applies the same prompted generation process incrementally:
\begin{equation}
\Delta_t = \text{Model}_{g}(\mathcal{S}_{t-1}, \tau_t, r_t), \quad \mathcal{S}_t = \text{Apply}(\Delta_t, \mathcal{S}_{t-1}).
\end{equation}
The patch $\Delta_t$ may add, revise, or delete skills. This incremental update preserves the skill evolution history and avoids regenerating the entire library from scratch.

\textbf{Family reset.} This design follows the goal of \benchname{} to evaluate lifelong learning within a single class of agent tasks rather than the agent's ability to apply skills precisely across heterogeneous workflows. Interleaving tasks from different workflows in a single evaluation stream would introduce unnecessary system noise, including confounds from skill retrieval mechanisms.

\paragraph{Skill patch schema.}
We use a minimal auditable interface for skill evolution to avoid complex instruction-following effects that could compromise evaluation fairness. Skill patch generation is driven by the model's native capability and is decoupled from the agent harness used for task execution. In the equations above, $g$ denotes the fixed prompt template that specifies how the model should generate skill patches; this template is provided in Appendix~\ref{appendix:skill-prompt-template}. Each patch contains three fields: \texttt{summary}, \texttt{upsert\_files}, and \texttt{delete\_paths}. Together, these fields are sufficient to record the high-level lesson, create or overwrite \texttt{SKILL.md} files and helper scripts, and remove obsolete artifacts. The resulting skills follow the same standard skill structure described above. This file-level interface preserves patch history, supports update tracking over time, and makes failure modes such as uncontrolled skill growth directly inspectable when redundant or low-value artifacts accumulate in the library.

\paragraph{Skill use detection.}
Each task is completed under the planning policy of its native agent harness. Because these harnesses already support skills, the model can choose to use skills natively without any additional control mechanism. We only extract task-completion outcomes and execution-trace events associated with library files, such as skill reads or skill calls, to detect skill use and compute the skill reuse rate.

This protocol evaluates not only whether an agent can benefit from a static external memory, but also whether it can continuously maintain, repair, and improve a reusable skill repository over time.

\subsection{Metrics}

We evaluate agents using three metric groups:
\begin{itemize}
    \item \textbf{Task success rate}: whether the final output satisfies the task verifier.
    \item \textbf{Efficiency}: runtime cost per task, reported through interaction turns, monetary cost, and output tokens.
    \item \textbf{Skill generation and reuse}: statistics that record both the average number of skills retained in the final family-local library and the rate at which previously stored skills are reused in later tasks.
\end{itemize}
\section{Experiments}

\subsection{Experimental Setup}

We evaluate \benchname{} under the \textbf{Agentic Lifelong Learning} protocol. Within each task family, an agent begins with an empty skill library, solves the first task without prior skills, and then updates the library iteratively through \textit{skill patches} derived from execution trajectories and rubric feedback. To place each model in its strongest practical setting, we pair it with a matched agent wrapper and execution harness. We standardize only the interface for skill-patch generation by requiring a single-turn output format. This keeps the representation of generated skills consistent while preserving compatibility with the execution environment of each model.

The main experiments use Claude Code~\cite{anthropic2025claudecode}, Codex CLI~\cite{openai2025codexcli}, Qwen-Coder~\cite{qwen_code}, and Kimi-CLI~\cite{kimi_cli} as evaluation harnesses. Across these harnesses, we evaluate 11 model variants: Claude Sonnet 4.5, Claude Opus 4.5, Claude Sonnet 4.6, Claude Opus 4.6, MiniMax M2.5, MiniMax M2.7, GPT 5.4, GPT 5.3 Codex, Qwen-Coder-Next, Qwen3-Coder-480B, and Kimi K2.5.

\nolinenumbers
\begin{table*}[t]
    \centering
    \small
    \renewcommand{\arraystretch}{1.2} % 增加行间距，防止文字太挤
    \setlength{\tabcolsep}{4.5pt}    % 稍微调整列间距
    \caption{Main Experimental Results. All values are benchmark-level averages. \%comp. is task completion rate; Turns, Cost, and Out Tok. are per-task averages of interaction turns, USD cost, and output tokens (thousands). \#Skills is the cumulative number of skills generated within a task family, and \%use is the percentage of tasks that read or call at least one stored skill. In the $\Delta$ block, \%comp. and Turns are absolute changes, whereas \%Cost and \%Out Tok. are relative percentage changes. \colorbox{bettermore}{Green} indicates higher-is-better metrics, \colorbox{betterless}{blue} indicates lower-is-better metrics, and \textbf{bold} marks the best value in each column.} 
    \label{tab:main-exp}
    \resizebox{\textwidth}{!}{%
    % 修改列定义：全居中 'c'，并在指定位置加竖线 '|'
    \begin{tabular}{cc | cccc | cccccc | >{\columncolor{bettermore}}c >{\columncolor{betterless}}c >{\columncolor{betterless}}c >{\columncolor{betterless}}c}
        \toprule
        \multirow{2}{*}{\textbf{Agent}} & \multirow{2}{*}{\textbf{Model}} & \multicolumn{4}{c|}{\textbf{vanilla}} & \multicolumn{6}{c|}{\textbf{skills evolve}} & \multicolumn{4}{c}{$\boldsymbol{\Delta}$} \\
        \cmidrule(lr){3-6} \cmidrule(lr){7-12} \cmidrule(lr){13-16}
        & & \%comp.$\uparrow$ & Turns$\downarrow$ & Cost(USD)$\downarrow$ & Out Tok.(K)$\downarrow$ & \%comp.$\uparrow$ & Turns$\downarrow$ & Cost$\downarrow$ & Out Tok.(K)$\downarrow$ & \#Skills & \%use & \%comp.$\uparrow$ & Turns$\downarrow$ & \%Cost$\downarrow$ & \%Out Tok.$\downarrow$ \\
        \midrule
        \multirow{6}{*}{Claude Code} 
            & Claude Sonnet 4.5 & 49.4 & 25.04 & 0.293 & 1.07 & 55.42 & 24.88 & 0.246 & 0.85 & 2.55 & 72.89 & +6.02 & -0.16 & -16.04 & \textbf{-20.56} \\
            & Claude Opus 4.5 & 58.43 & 18.83 & 0.571 & 1.5 & 60.84 & 18.31 & 0.384 & 1.4 & 1.5 & 60.84 & +2.41 & -0.52 & \textbf{-32.87} & -6.67 \\
            & Claude Sonnet 4.6 & 56.63 & 17.48 & 0.168 & 1.23 & 56.63 & 17.42 & 0.245 & 1.59 & 2.55 & 53.01 & +0.00 & -0.06 & +45.83 & +29.27 \\
            & Claude Opus 4.6 & \textbf{62.65} & 17.34 & 0.665 & 3.00 & \textbf{71.08} & 19.00 & 0.615 & 2.39 & 1.05 & 45.78 & \textbf{+8.43} & 1.66 & -7.52 & -20.33 \\
            & MiniMax M2.5 & 28.31 & 35.22 & \textbf{0.010} & \textbf{0.44} & 34.94 & 34.01 & \textbf{0.010} & \textbf{0.54} & 2.50 & 32.53 & +6.63 & \textbf{-1.21} & 0 & +22.73 \\
            & MiniMax M2.7 & 37.35 & 25.44 & 0.012 & 0.5 & 36.75 & 27.42 & 0.017 & 0.96 & 4.6 & 51.2 & -0.6 & +1.98 & +41.67 & +92 \\
        \midrule
        \multirow{3}{*}{Codex CLI}
            & GPT 5.4 & 33.13 & 23.89 & 0.41 & 4.05 & 36.75 & 24.17 & 0.459 & 4.43 & 1.05 & 81.33 & +3.62 & +0.28 & +11.95 & +9.38 \\
            & GPT 5.3 Codex & 52.41 & 17.74 & 0.492 & 6.8 & 46.39 & 17.14 & 0.434 & 6.82 & 1.1 & 84.94 & -6.02 & -0.6 & -11.79 & +0.29 \\
        \midrule
        \multirow{2}{*}{Qwen Coder}
            & Qwen-Coder-Next & 45.18 & 18.64 & 0.103& 9.74 & 44.58 & 19.91 & 0.113 & 10.69 & 5.45 & 12.05 & -0.60 & +1.27 & +9.71 & 9.75 \\
            & Qwen3-Coder-480B & 24.7 & 26.22 & 0.189 & 12.58 & 24.1 & 28.8 & 0.199 & 12.12 & 5.2 & 66.87 & -0.6 & +2.58 & +5.29 & -3.66 \\
        \midrule
        \multirow{1}{*}{Kimi CLI}
            & Kimi K2.5 & 55.42 & \textbf{12.62} & 0.103 & 7.31 & 56.02 & \textbf{11.51} & 0.104 & 7.10 & 1.50 & 66.87 & +0.60 & -1.11 & +0.97 & -2.87 \\
        \bottomrule
    \end{tabular}%
    }
\end{table*}

\begin{figure*}[t]
    \centering
    \begin{minipage}[t]{0.48\textwidth}
        \centering
        \includegraphics[width=0.94\linewidth]{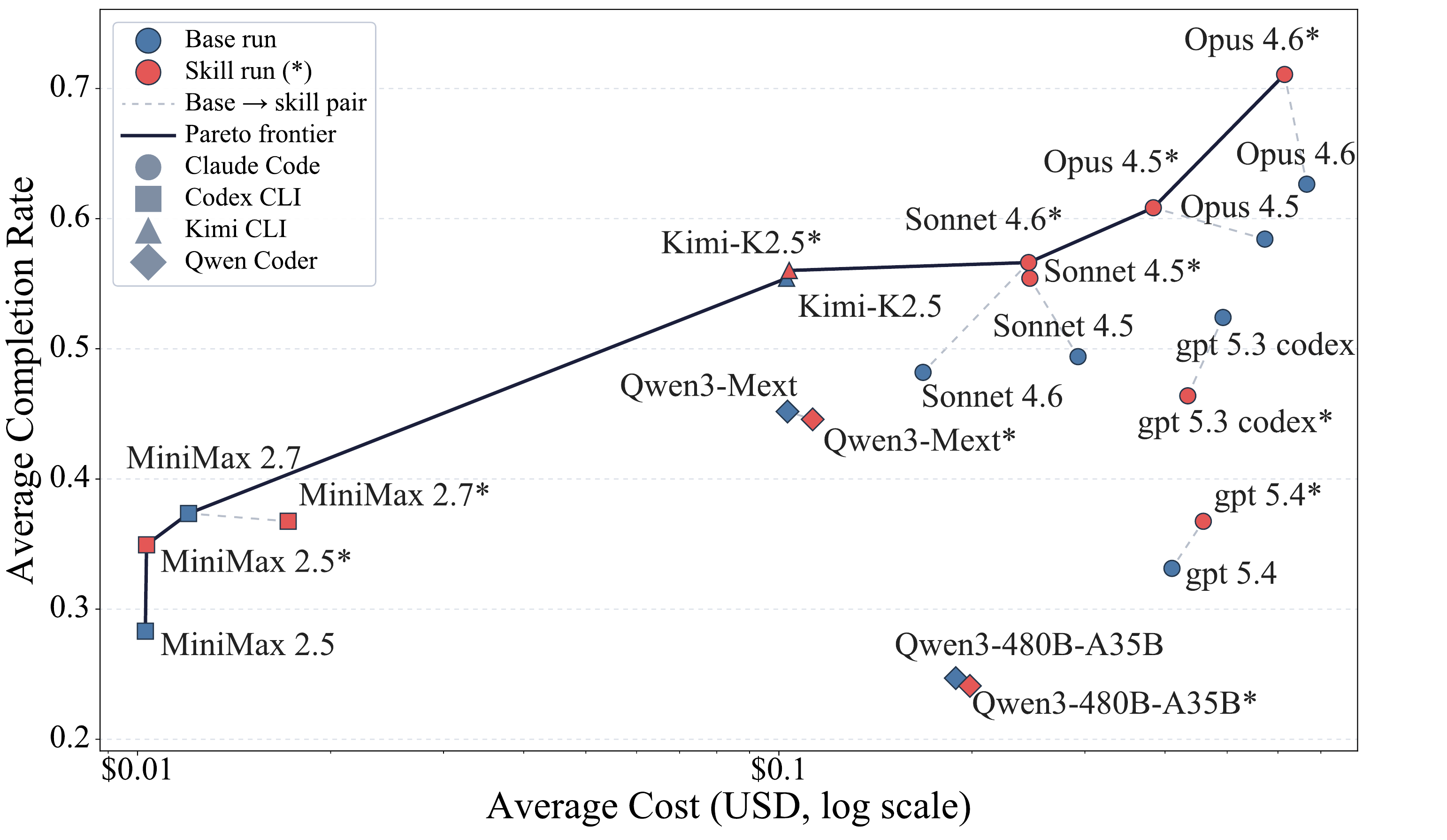}
        \captionof{figure}{Completion--Cost Pareto Frontier. Each point is one evaluated agent-model setting under vanilla execution or lifelong skill evolution. Some settings shift toward higher completion with comparable or lower cost, while others gain little or regress despite additional spending.}
        \label{fig:pareto-cost}
    \end{minipage}\hfill
    \begin{minipage}[t]{0.48\textwidth}
        \centering
        \includegraphics[width=\linewidth]{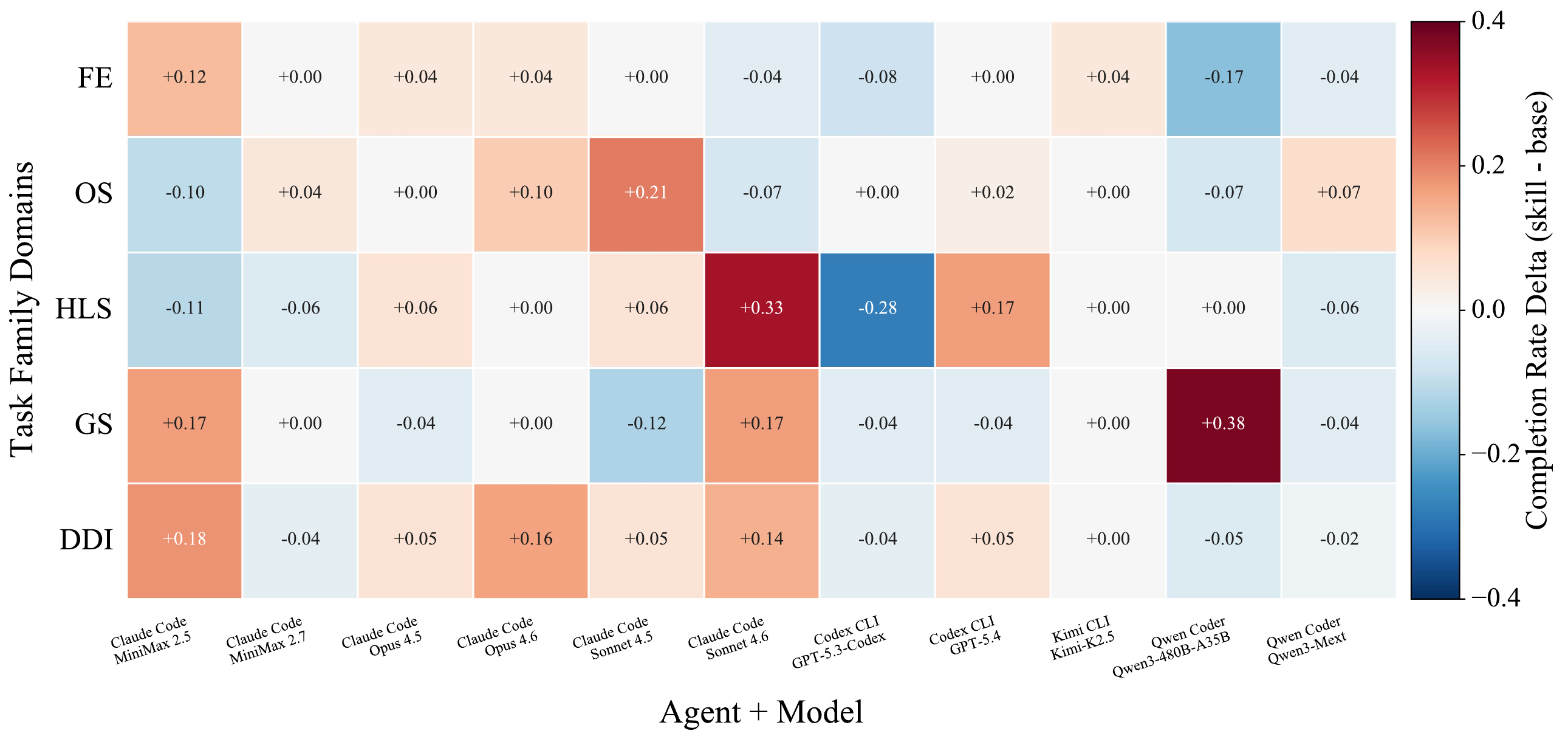}
        \captionof{figure}{Domain-Grouped Completion Gains from Skill Evolution. Rows correspond to evaluated agent-model settings, while columns are organized into coarse benchmark groups. Abbreviations: FE = Finance \& Economics; OS = Operations \& Supply Chain; HLS = Healthcare \& Life Sciences; GS = Governance \& Strategy; DDI = Data \& Document Intelligence.}
        \label{fig:skill-gain-heatmap}
    \end{minipage}
\end{figure*}

\subsection{Observations}

Table~\ref{tab:main-exp} shows that skill evolution yields substantial gains for some models, but the effect varies sharply across settings. The clearest positive case is Claude Opus 4.6, which increases completed tasks from 104 to 118 out of 166 and raises mean task success from 62.65\% to 71.08\% (+8.43 points). Positive transfer also appears for MiniMax M2.5, which improves from 28.31\% to 34.94\% (+6.63 points), for Claude Sonnet 4.5, which improves from 49.40\% to 55.42\% (+6.02 points), for GPT 5.4, which rises from 33.13\% to 36.75\% (+3.62 points), for Claude Opus 4.5, which rises from 58.43\% to 60.84\% (+2.41 points), and for Kimi K2.5, which increases modestly from 55.42\% to 56.02\% (+0.60 points).

By contrast, Table~\ref{tab:main-exp} also contains several settings in which skill evolution does not help or directly reduces performance. GPT 5.3 Codex declines from 52.41\% to 46.39\% (-6.02 points). Qwen-Coder-Next declines from 45.18\% to 44.58\% (-0.60 points), Qwen3-Coder-480B declines from 24.70\% to 24.10\% (-0.60 points), and MiniMax M2.7 declines from 37.35\% to 36.75\% (-0.60 points). Claude Sonnet 4.6 remains unchanged at 56.63\% under both settings.

Figure~\ref{fig:pareto-cost} provides a complementary efficiency view of the same settings. Several points move toward higher completion with comparable or lower cost, whereas others move toward higher cost with limited gains or no gains. Figure~\ref{fig:skill-gain-heatmap} shows that gains are broadly distributed across task-family groups rather than concentrated in a single category, which suggests that current models do not yet exhibit a clear preference for any particular task type during skill evolution. At the same time, some group-level differences remain: Finance \& Economics contains more negative gains, whereas Data \& Document Intelligence more often shows positive transfer. Appendix Figure~\ref{fig:skill-gain-heatmap-all} provides the full family-level heatmap for a more detailed view.

\subsection{Main Results}

To better understand model behavior under the Agentic Lifelong Learning protocol, we analyze the full set of experimental results from all 11 evaluated models and summarize the following findings.

\noindent\textbf{Finding 1: Opus 4.6 comes closest to stable skill-based learning.} Among the evaluated models, Opus 4.6 provides the clearest evidence that skill revision can improve the procedure stored in the library rather than merely accumulate successful traces. Table~\ref{tab:main-exp} and Figure~\ref{fig:pareto-cost} show the strongest overall improvement for this setting, with higher completion at lower cost. Trace analysis further shows cases in which repaired skills are reused after concrete failures, including spreadsheet workflows that require cached-value handling. We also conduct a control experiment that simply prepends the full prior interaction history as additional context. On Claude Opus 4.6, this setting reaches only 51.04\%, below both vanilla and the full protocol (Appendix Table~\ref{tab:history-context-control}), which suggests that the gain is not explained by longer raw context alone. Within the scope of our benchmark, this is the clearest instance of stable library-level improvement.

\noindent\textbf{Finding 2: Incorrect skills create systematic downstream drift.} Once an incorrect skill is written into the library, later tasks often inherit the same flawed abstraction, turning a local error into a sequence-level pattern. This pattern likely reflects the difficulty of escaping a self-generated logic distribution, which suggests that, while external skills can amplify capability, they can also amplify error.

\noindent\textbf{Finding 3: Unified high-utility skills outperform fragmented task-specific skills.} The strongest libraries are usually organized around one or a few reusable skills that are revised repeatedly as new variants appear. In Table~\ref{tab:main-exp}, stronger settings tend to end with smaller final skill inventories, whereas weaker settings often accumulate more skills without comparable gains. This pattern suggests that consolidation is generally more valuable than proliferation.

\noindent\textbf{Finding 4: Qwen and part of MiniMax mainly fail through skill inflation.} For Qwen and some MiniMax settings, the main problem is often not skill absence but uncontrolled growth of overlapping skills. Appendix Figure~\ref{fig:skill-count-growth} shows that the accumulated number of stored skills in these settings increases almost monotonically with task index, which suggests that they tend to summarize nearly every task into an additional skill. Yet Table~\ref{tab:main-exp} still shows weak or negative benchmark-level gains, and Figure~\ref{fig:skill-gain-heatmap} does not show correspondingly broad improvement across task families. In these cases, the dominant failure mode is better described as fragmentation through skill inflation than as skill scarcity.

\noindent\textbf{Finding 5: Codex is relatively strong at consolidating variants into one evolving core skill.} Codex shows a clearer tendency than most models to absorb nearby task variants into a shared evolving skill rather than proliferate many narrowly specialized entries. Appendix Figure~\ref{fig:skill-file-kind-composition} shows that Codex-based libraries remain relatively compact in file count compared with many other settings. However, Table~\ref{tab:main-exp} also shows that this compactness does not translate into stronger task completion: compared with Opus settings, which also keep their libraries relatively small, Codex does not achieve comparable end-to-end gains under the full harness.

\noindent\textbf{Finding 6: The key model gap lies in repairing bad skills, not in writing skills.} Most models can write some skill after completing or attempting a task, so raw skill generation is not the main source of variation. The larger difference is whether a model can recognize an incorrect skill, revise it, and obtain better behavior on later tasks. Across the table, the frontier plot, and the traces, positive transfer aligns more closely with effective skill repair than with skill volume alone.

\noindent\textbf{Summary.}

These findings show that the effects of iterative skill evolution depend strongly on model capability. Across settings, compact and well-revised skills are consistently more effective than fragmented, isolated skill entries, while incorrect early skills can induce persistent negative transfer and substantially degrade later performance. More broadly, the practical value of skill evolution appears to depend less on writing more skills than on maintaining a small, repairable library of high-utility procedures. This property may also make it easier to retrieve relevant skills when extending such systems to more open-ended environments.

\section{Related Work}

\subsection{Execution Environments and Benchmarks}
A growing line of work evaluates agent capabilities in controlled environments with tool use and multi-step reasoning~\cite{patil2025berkeley,liu2023agentbench,trivedi2024appworld}. 
Some benchmarks further focus on reliability aspects such as error correction in tool calling~\cite{huang-etal-2025-critictool}, as well as the inefficiency patterns that emerge during complex tool-integrated reasoning~\cite{su2026accuracyunveilinginefficiencypatterns}. 
Other work studies whether explicit skill usage improves performance in realistic software engineering settings~\cite{han2026swe}. 
Recent coding-agent benchmarks emphasize realistic, long-horizon tasks under shared Harbor-based execution setups for reproducibility and comparability~\cite{merrill2026terminal,xia2024top,deng2025swe,chou2025autocodebench,chen2026skillcraft}.

\subsection{Skills as Procedural Knowledge for Agents}
Recent studies treat \textbf{skills} as reusable procedural knowledge bridging models and workflows, including large-scale skill management, skill-aware benchmarking, and trajectory distillation into reusable skills~\cite{liang2026skillnet,pinchbench2026,ni2026trace2skill}. 
However, these works mainly focus on infrastructure or downstream performance, with limited evaluation of skill derivation and cross-task transfer.

\subsection{Automatic Skill Discovery and Evolution}
Another line of work explores automatic skill discovery and evolution from interaction, including distilling interaction patterns, refining skills through feedback and failures~\cite{zheng2025skillweaver,yang2026autoskill,alzubi2026evoskill,zhang2026memskill,wei2025evo,huang2026internalizing}, and self-improving models via self-generated supervision~\cite{han2026unicorn}. 
Some approaches further model skills as persistent and improvable structures or leverage long interaction traces for accumulation~\cite{xu2023xskill,zhou2026memento,xia2026skillrl,cao2026coding}. 
These methods demonstrate performance gains from skill accumulation, but focus more on optimization than on evaluating skill abstraction, revision, and transfer.
\section{Conclusion}

We introduce \benchname{}, a benchmark evaluating lifelong skill discovery and evolution in autonomous agents. It contributes a dual-agent construction pipeline, an explicit sequential evaluation protocol, and a task set organized around Domain-Agnostic Execution Flow to test skill transfer at the workflow level instead of superficial lexical overlap. Experiments show that lifelong skill evolution yields selective rather than universal gains: stronger agent--model stacks convert externalized experience into compact reusable procedures, whereas weaker ones often exhibit a creation--reuse coordination gap, fragmented skill libraries, and unstable reinforcement under feedback. This positions \benchname{} as both a benchmark release and an initial empirical characterization of skill evolution under one concrete external-memory mechanism.

More broadly, as autonomous agents become increasingly general-purpose, continuous skill evolution may provide a practical mechanism for acquiring and consolidating knowledge across domains. In this sense, enabling agents to revise skills from experience appears important for building more robust and adaptive lifelong learning systems.

\newpage
\bibliographystyle{plainnat}
\bibliography{reference}
\newpage
\appendix
\onecolumn

\section{Benchmark Details}
\subsection{Comparison}

\begin{table*}[htbp]
\centering
\scriptsize
\setlength{\tabcolsep}{4pt}
\caption{
Comparison between {\benchname} and representative benchmarks from the perspective of skill lifecycle evaluation.
\textbf{Skill Eval} denotes whether the benchmark explicitly evaluates the impact of skills;
\textbf{Self-Gen} indicates support for self-generated skills;
\textbf{Revision} refers to iterative updates or patching of skills;
\textbf{Lifelong} denotes sequential accumulation and reuse of skills across tasks;
\textbf{Transfer} indicates evaluation of cross-scenario procedural transfer;
\textbf{Traj-Grounded} denotes whether skills are extracted from interaction trajectories;
\textbf{Usage Eval} indicates whether the benchmark evaluates the alignment between skill utility and actual usage.
}
\label{table:compare_benchmarks}
\begin{tabular}{lccccccc}
\toprule
Benchmark & Skill Eval & Self-Gen & Revision & Lifelong & Transfer & Traj-Grounded & Usage Eval \\
\midrule
\textbf{Terminal-Bench}~\citep{merrill2026terminal} & \xmark & \xmark & \xmark & \xmark & \xmark & \xmark & \xmark \\
\textbf{SWE-bench}~\citep{jimenez2023swe} & \xmark & \xmark & \xmark & \xmark & \xmark & \xmark & \xmark \\
\textbf{SkillsBench}~\citep{li2026skillsbench} & \cmark & \cmark & \xmark & \xmark & \xmark & \xmark & \xmark \\
\textbf{PinchBench}~\citep{pinchbench2026} & \xmark & \xmark & \xmark & \xmark & \xmark & \xmark & \xmark \\
\textbf{SWE-Skills-Bench}~\citep{han2026swe} & \cmark & \xmark & \xmark & \xmark & \xmark & \xmark & \xmark \\
\textbf{\benchname\ (Ours)} & \cmark & \cmark & \cmark & \cmark & \cmark & \cmark & \cmark \\
\bottomrule
\end{tabular}
\end{table*}

\subsection{Benchmark Statistics}

The final benchmark contains 20 task families and 166 tasks organized under five broad domains: \textit{Finance \& Economics}, \textit{Operations \& Supply Chain}, \textit{Healthcare \& Life Sciences}, \textit{Governance \& Strategy}, and \textit{Data \& Document Intelligence}. These five domains cover 20 workflow families in total, with family sizes ranging from 8 to 9 tasks. During construction, more than 200 candidate tasks are generated. Thirty-three of them are filtered out before final inclusion because of environment failures, ambiguity, invalid difficulty ordering, or workflow inconsistency.

\subsection{Workflow Families}

This section lists the 20 workflow families by their \emph{workflow-definition names} rather than internal source identifiers. The grouping is derived from the curated benchmark spreadsheet and uses only the formal family names reported in the main paper.

\begin{itemize}
    \item \textbf{Finance \& Economics}: \textit{Industry Correlation Analysis}, \textit{Financial Statement Rolling}, and \textit{SEC 13F Financial Analysis}.
    \item \textbf{Operations \& Supply Chain}: \textit{Supply Chain Replenishment}, \textit{Production Capacity Planning}, \textit{Inventory \& Finance Integration}, \textit{DMAIC Quality Analysis}, and \textit{Operational Recovery Planning}.
    \item \textbf{Healthcare \& Life Sciences}: \textit{Healthcare Cost-Benefit Analysis} and \textit{Medical Data Standardization}.
    \item \textbf{Governance \& Strategy}: \textit{Distribution Center Auditing}, \textit{Compensation Scenario Modeling}, and \textit{Document Fraud Detection}.
    \item \textbf{Data \& Document Intelligence}: \textit{Embedded Data Repair}, \textit{OCR Data Extraction}, \textit{HWPX Document Automation}, \textit{Cross-Format Data Reconciliation}, \textit{Weighted Risk Assessment}, \textit{PPT Formatting Optimization}, and \textit{Sales Pivot Analysis}.
\end{itemize}

\subsection{Workflow Details}

Table~\ref{tab:workflow-details} provides a compact view of the 20 workflow families. Internal source identifiers are omitted. Instead, the table reports the workflow definition, seed-benchmark lineage, a short task summary, and the corresponding DAEF description. The \textbf{Source} column indicates whether the family is derived from a seed task selected from SkillsBench or GDPval.

\begin{longtable}{p{0.16\textwidth}p{0.12\textwidth}p{0.25\textwidth}p{0.35\textwidth}}
    \caption{Workflow Families and Seed-Benchmark Lineage} \label{tab:workflow-details} \\
    \toprule
    \textbf{Workflow definition} & \textbf{Source} & \textbf{Task summary} & \textbf{DAEF} \\
    \midrule
    \endfirsthead
    \toprule
    \textbf{Workflow definition} & \textbf{Source} & \textbf{Task summary} & \textbf{DAEF} \\
    \midrule
    \endhead
    \textit{Industry Correlation Analysis} & SkillsBench & Detrend revenue series for paired industries, compute Pearson correlations, and judge cyclical association. & read source series $\rightarrow$ extract target fields $\rightarrow$ normalize and align time series $\rightarrow$ compute annualized and deflated values $\rightarrow$ compute filtered correlation signals $\rightarrow$ compare cyclical association $\rightarrow$ output scalar result. \\
    \textit{Financial Statement Rolling} & GDPval & Update monthly financial workbooks with reserve merges, accrual rolling, deferred revenue, liabilities, warranties, project cost, rebates, and commissions. & read multi-source ledgers $\rightarrow$ extract target entries $\rightarrow$ filter duplicate and invalid rows $\rightarrow$ align override and adjustment items $\rightarrow$ compute rolling balances $\rightarrow$ compare against control totals $\rightarrow$ update and output summary workbook. \\
    \textit{SEC 13F Financial Analysis} & SkillsBench & Analyze quarterly SEC 13F filings for cross-period reconciliation, fund categorization, holding changes, issuer ownership, and manager comparison. & read multi-period filings $\rightarrow$ extract entity records $\rightarrow$ normalize identifiers $\rightarrow$ align cross-period holdings $\rightarrow$ compute target aggregates $\rightarrow$ compare across periods $\rightarrow$ output structured analysis. \\
    \textit{Supply Chain Replenishment} & GDPval & Produce replenishment plans across categories such as bakery, beauty, frozen meals, household routes, pet care, produce, and snacks. & read inventory and transit data $\rightarrow$ extract target SKUs $\rightarrow$ filter duplicate and invalid rows $\rightarrow$ align master data with arrivals $\rightarrow$ compute coverage and shortage timing $\rightarrow$ compare against replenishment thresholds $\rightarrow$ output detail and action sheets. \\
    \textit{Production Capacity Planning} & GDPval & Build and vary a manufacturing catch-up capacity plan under demand, backlog, overtime, and buffer constraints. & read demand and parameters $\rightarrow$ retrieve initial state $\rightarrow$ compute period-level demand and backlog updates $\rightarrow$ compare candidate resource actions $\rightarrow$ compute overtime and buffer outcomes $\rightarrow$ output planning workbook. \\
    \textit{Inventory \& Finance Integration} & GDPval & Integrate replenishment planning with financial rolling scenarios across mixed business contexts. & read multi-source data $\rightarrow$ extract and normalize keys and time windows $\rightarrow$ align inventory and finance records $\rightarrow$ compute rolling deltas and coverage $\rightarrow$ compare against exception rules $\rightarrow$ output structured results. \\
    \textit{DMAIC Quality Analysis} & GDPval & Perform DMAIC Analyze-stage diagnostics across logistics, hospitals, SOC alerts, university IT, DevOps, and field service settings. & read process data $\rightarrow$ extract analysis window $\rightarrow$ filter valid samples $\rightarrow$ compute grouped statistics $\rightarrow$ compare trend and significance signals $\rightarrow$ detect anomalies $\rightarrow$ output structured diagnostics. \\
    \textit{Operational Recovery Planning} & GDPval & Create recovery plans under multi-scenario operational disruptions such as harvesting, data centers, warehouses, manufacturing, radiology, and returns. & read baselines and constraints $\rightarrow$ extract planning horizon $\rightarrow$ compute candidate recovery plans $\rightarrow$ compare capacity and calendar outcomes $\rightarrow$ detect backlog gaps $\rightarrow$ output plan sheets and summaries. \\
    \textit{Healthcare Cost-Benefit Analysis} & GDPval & Compare healthcare supply-chain scenarios under different cycle and batch settings to assess cost-benefit trade-offs. & read inputs and mappings $\rightarrow$ extract and normalize aliases and keys $\rightarrow$ filter valid records and exclusions $\rightarrow$ align coverage with adjustments $\rightarrow$ compute revenue cost and margin scenarios $\rightarrow$ compare trade-off thresholds $\rightarrow$ output structured conclusions. \\
    \textit{Medical Data Standardization} & SkillsBench & Standardize medical laboratory data across units, precision rules, and formatting conventions. & read raw tables and templates $\rightarrow$ extract valid rows $\rightarrow$ align with target schema $\rightarrow$ normalize numeric formats and units $\rightarrow$ detect anomalies or missing values $\rightarrow$ validate required precision $\rightarrow$ output standardized file. \\
    \textit{Distribution Center Auditing} & GDPval & Audit distribution-center records for inventory discrepancies, outbound lists, receiving anomalies, return handling, SLA compliance, labor, and trailer detention. & read source data $\rightarrow$ extract audit fields $\rightarrow$ compute row-wise rule checks $\rightarrow$ detect violations and anomalies $\rightarrow$ compare totals across dimensions $\rightarrow$ output audit artifacts. \\
    \textit{Compensation Scenario Modeling} & GDPval & Build and refresh multi-year compensation models for organizations such as orchestras, universities, property firms, airlines, and construction teams. & read assumptions and master data $\rightarrow$ retrieve model structure and references $\rightarrow$ align base records across sheets $\rightarrow$ compute period-wise compensation values $\rightarrow$ update and compare scenario indicators $\rightarrow$ output completed model. \\
    \textit{Document Fraud Detection} & SkillsBench & Review commercial documents across clinics, trials, maintenance logs, travel, and cold-chain records to detect inconsistencies and suspicious items. & read source documents $\rightarrow$ extract candidate records and references $\rightarrow$ normalize names and identifiers $\rightarrow$ align cross-document evidence $\rightarrow$ compare claimed facts with authorizations $\rightarrow$ detect anomaly types $\rightarrow$ output structured flags. \\
    \textit{Embedded Data Repair} & SkillsBench & Repair or refresh embedded spreadsheet objects inside PowerPoint files for domains such as chemistry, catalysts, or foreign exchange. & read document container $\rightarrow$ extract embedded tables and target cells $\rightarrow$ retrieve label mappings $\rightarrow$ update cell values and formulas $\rightarrow$ validate dependencies $\rightarrow$ output repaired document. \\
    \textit{OCR Data Extraction} & SkillsBench & Extract structured records from scanned images such as legal settlements, measurements, orders, invoices, fuel slips, and medicine labels. & read scanned images $\rightarrow$ extract OCR fields and target records $\rightarrow$ normalize dates amounts and codes $\rightarrow$ align with template rows $\rightarrow$ validate supplemented references $\rightarrow$ output structured results. \\
    \textit{HWPX Document Automation} & SkillsBench & Fill Korean HWPX templates for clinics, announcements, inventory, proposals, renewals, audits, vendor directories, and training feedback. & read templates and data sources $\rightarrow$ extract placeholders and values $\rightarrow$ normalize transformed values and enumerations $\rightarrow$ update filled content and structure $\rightarrow$ validate document integrity $\rightarrow$ output completed file. \\
    \textit{Cross-Format Data Reconciliation} & SkillsBench & Compare archived PDF snapshots with current Excel data to find added, deleted, or modified records across domains. & read archived snapshot and current spreadsheet $\rightarrow$ extract primary keys and records $\rightarrow$ normalize and align across formats $\rightarrow$ compare additions deletions and modifications $\rightarrow$ detect field-level differences $\rightarrow$ output structured results. \\
    \textit{Weighted Risk Assessment} & SkillsBench & Implement weighted risk formulas in Excel for scenarios such as API SLA, campus budgeting, factories, hospitals, energy, cloud reliability, and ports. & read workbook and lookup tables $\rightarrow$ retrieve input and output ranges $\rightarrow$ update formulas with target cells $\rightarrow$ compute weighted indicators and risk scores $\rightarrow$ validate workbook consistency $\rightarrow$ output result workbook. \\
    \textit{PPT Formatting Optimization} & SkillsBench & Normalize image-title formatting in PowerPoint decks by unifying fonts, positions, and index pages. & read presentation objects $\rightarrow$ extract target titles and tags $\rightarrow$ normalize text font and layout attributes $\rightarrow$ update positions and alignment $\rightarrow$ validate summary-page consistency $\rightarrow$ output formatted document. \\
    \textit{Sales Pivot Analysis} & SkillsBench & Build pivot-style analysis from PDF catalogs and Excel transaction data for sales, budgets, payroll, registration, inventory, libraries, quality, and grades. & read multi-source data $\rightarrow$ extract and normalize fields and dimensions $\rightarrow$ align supplementary joins $\rightarrow$ compute pivot indicators $\rightarrow$ compare grouped summaries $\rightarrow$ output report workbook. \\
    \bottomrule
\end{longtable}

\subsection{DAEF Node Vocabulary}
\label{sec:appendix-daef-vocab}

Table~\ref{tab:daef-vocab} lists the controlled node labels used when annotating seed-task workflow graphs. The vocabulary is intentionally compact and coarse-grained so that it captures transferable procedural roles rather than low-level commands or domain-specific business concepts.

\begin{table}[t]
    \centering
    \small
    \begin{tabular}{p{0.12\textwidth}p{0.23\textwidth}p{0.53\textwidth}}
        \toprule
        \textbf{Node label} & \textbf{Operational meaning} & \textbf{Typical examples in released workflows} \\
        \midrule
        read & Load or inspect source files, tables, documents, images, or structured records & Read ledgers, filings, scanned images, templates, workbooks, PPTX containers, and PDF snapshots \\
        extract & Identify and isolate target fields, records, entities, or patterns from complex or unstructured sources & Extract OCR fields from scanned images, extract candidate records from documents, extract embedded tables from containers \\
        retrieve & Locate or select target states, ranges, placeholders, or action candidates from available context & Retrieve initial state, retrieve target entries, retrieve audit fields, retrieve target cells, retrieve placeholders, retrieve input/output ranges \\
        normalize & Standardize schema, keys, units, formatting, aliases, or transformed values before downstream reasoning & Normalize entity identifiers, numeric formats, lab units, dates, codes, font attributes, and transformed enumerations \\
        filter & Retain valid records and remove irrelevant, duplicate, malformed, or excluded cases & Filter duplicate ledger rows, filter excluded healthcare cases, filter invalid samples, filter malformed source entries \\
        align & Join, reconcile, or map multiple sources under a shared key, structure, or layout & Align time series, align cross-period holdings, align inventory with arrivals, align finance records, align label mappings, align records across formats \\
        compute & Derive intermediate or final quantities through formulas, aggregation, scoring, iterative updates, or planning operations & Compute rolling balances, compute coverage windows, compute backlog updates, compute revenue and margin scenarios, compute weighted risk scores, compute pivot indicators \\
        compare & Contrast values, scenarios, plans, grouped summaries, or cross-period states against a threshold or reference target & Compare cyclical association, compare control totals, compare replenishment thresholds, compare candidate resource actions, compare trade-off thresholds, compare grouped summaries, compare record changes \\
        detect & Identify anomalies, violations, exceptions, gaps, or field-level changes that require action & Detect suspicious claims, detect audit violations, detect missing values, detect backlog gaps, detect anomaly types, detect field-level differences \\
        update & Modify, fill, repair, or refresh existing artifacts such as cells, documents, or configurations & Update embedded cell values, update workbook formulas, update document formatting, update template fields, update configuration values \\
        validate & Check whether outputs satisfy structural, numeric, dependency, formatting, or integrity constraints before release & Validate required precision, validate workbook consistency, validate formulas and dependencies, validate cache integrity, validate reference supplements, validate summary-page consistency \\
        output & Write the final artifact in the required structure, file format, workbook, document, or report layout & Output scalar results, output structured diagnostics, output linked summary workbooks, output audit artifacts, output repaired documents, output formatted presentations \\
        \bottomrule
    \end{tabular}
    \caption{Controlled DAEF Node Vocabulary}
    \label{tab:daef-vocab}
\end{table}

\paragraph{DAEF Annotation Rules.}
The annotation process follows five rules. First, each task is annotated with 5--8 meta-step nodes. Second, node labels must be selected only from the controlled vocabulary in Table~\ref{tab:daef-vocab}, and annotators are not allowed to invent new labels. Third, dependency edges are directed from prerequisite operations to subsequent operations, and cycles are not permitted. Fourth, when two operations can be executed in parallel, both are connected to the same downstream node. Fifth, after annotation, annotators must provide a brief operational rationale in one sentence explaining why each node label is selected.

\paragraph{DAEF Agreement Screening.}
To verify that a seed task admits a stable workflow abstraction, two expert annotators independently produce DAEF annotations for each candidate task in the final screening pool, which yields 30 retained tasks after screening. The annotators then score the other annotator's graph against their own judgment, checking agreement on the meta-step decomposition, node labels, and dependency structure. A candidate passes only when both annotators judge the two annotations to be mutually consistent. Otherwise, we treat the seed task as lacking a stable execution workflow and discard it from benchmark construction.

\subsection{Allowed Surface Variation Types}
\label{sec:appendix-daef-variation}

Table~\ref{tab:daef-variation} summarizes the variation dimensions that annotators may allow within a family while preserving a shared DAEF. These dimensions are intended to alter surface realization and difficulty without introducing a new workflow core.

\begin{table}[t]
    \centering
    \small
    \begin{tabular}{p{0.18\textwidth}p{0.26\textwidth}p{0.43\textwidth}}
        \toprule
        \textbf{Variation type} & \textbf{What may change} & \textbf{Representative examples in the release} \\
        \midrule
        Domain entities & Business objects, named fields, organizational units, or scenario semantics & Vendor vs. patient vs. warehouse records; holdings vs. inventory vs. clinical measurements \\
        File format & Surface container or artifact type while preserving the same procedural core & XLSX vs. CSV vs. JSON inputs; PPTX embedded spreadsheets vs. standalone workbooks; OCR image vs. PDF table \\
        Data scale & Number of rows, time periods, scenarios, or cross-file joins & More SKUs, longer time horizons, larger audit logs, additional quarterly filings \\
        Noise and quality & Missing values, alias ambiguity, OCR noise, inconsistent formatting, or extra distractors & Misspelled headers, low-quality scans, inconsistent units, partial records, cluttered slides \\
        Constraint regime & Thresholds, policy rules, reconciliation targets, or validation tolerances & Different SLA cutoffs, coverage rules, accounting constraints, exception thresholds \\
        Instruction surface & Wording, ordering, or presentation of the task request without changing the workflow core & Domain-specific phrasing, reformulated reporting goals, alternate output wording \\
        Difficulty level & The amount of bookkeeping, exception handling, or multi-step reasoning required & Additional scenario branches, more exception rules, harder reconciliation chains, denser formatting checks \\
        \bottomrule
    \end{tabular}
    \caption{Allowed Surface Variation Types Within a DAEF Family}
    \label{tab:daef-variation}
\end{table}

\subsection{Human Verification Checklist}

Human reviewers inspect each candidate family with the following checklist:
\begin{itemize}
    \item \textbf{Solvability}: a correct solution path exists under the provided environment and assets.
    \item \textbf{Logical consistency}: instructions, assets, and expected outputs do not contradict one another.
    \item \textbf{Environment closure}: the task can be completed with the provided files, tools, and runtime.
    \item \textbf{Difficulty ordering}: later tasks in a family are not easier than earlier tasks under the intended curriculum.
    \item \textbf{No skill leakage}: instructions do not reveal which exact skill should be used.
\end{itemize}

For family acceptance, reviewers additionally assess whether the candidate tasks satisfy the DAEF membership rule:
\begin{itemize}
    \item \textbf{Shared operation inventory}: the core operation types are consistent across tasks.
    \item \textbf{Shared dependency topology}: these operations follow the same dependency pattern rather than merely similar names.
    \item \textbf{Grounding variation only}: differences arise from domain entities, file formats, noise, scale, or instruction phrasing rather than from a new workflow core.
    \item \textbf{Transfer plausibility}: a procedure learned on earlier tasks should plausibly transfer to later tasks without requiring a new workflow.
\end{itemize}

\section{Skill Evolution Details}
\subsection{Skill Patch Format}

Each skill patch in the current framework is represented as a JSON object with three top-level fields:
\begin{itemize}
    \item \texttt{summary}: a natural-language description of the newly extracted lesson or repair.
    \item \texttt{upsert\_files}: a mapping from file paths to new or updated file contents.
    \item \texttt{delete\_paths}: a list of obsolete files to remove from the skill library.
\end{itemize}

This schema is used as a minimal auditable interface for file-level updates rather than as an optimal or exhaustive formalism for skill evolution. Its purpose is to make patch history inspectable, support update tracking over time, and make failure modes such as uncontrolled skill growth directly visible when redundant or low-value content accumulates.

\subsection{Skill Patch Generation Prompt Template}
\label{appendix:skill-prompt-template}

The following appendix presents the prompt template used for skill patch generation. For readability, we separate the system-level instruction from the user-level prompt template and render them as boxed templates.

\makeatletter
\@ifundefined{lstdefinestyle}{\RequirePackage{listings}}{}
\makeatother

\lstdefinestyle{prompttemplate}{
    basicstyle=\ttfamily\fontsize{7pt}{8pt}\selectfont,
    frame=none,
    breaklines=true,
    backgroundcolor=\color{white},
    breakatwhitespace=true,
    breakindent=0pt,
    numbers=none,
    numbersep=5pt,
    xleftmargin=5pt,
    aboveskip=2pt,
    belowskip=2pt,
}

\newtcolorbox{prompttemplatebox}[2][]{
  top=10pt,
  colback=white,
  enhanced,
  center,
  breakable,
  fonttitle=\bfseries,
  title=#2,
  #1
}

\begin{prompttemplatebox}[breakable]{System Prompt Template}
\begin{lstlisting}[style=prompttemplate]
A skill is a reusable capability package, not a task-specific note. Improve the library so a future agent can solve similar tasks faster, more reliably, and with fewer failed attempts.

Hard structural rules (treat these as requirements, not suggestions):

- A new skill should normally live in its own capability-named directory, for example `fill-pdf-forms/` or `api-debugging/`.

- Every skill directory must contain `SKILL.md` at its root.

- `SKILL.md` must begin at the first character of the file with YAML frontmatter in exactly this shape:

---
name: <skill-name>
description: <what the skill does and when to use it>
---

- The frontmatter may contain exactly two keys: `name` and `description`. Do not add any other metadata keys.

- After the closing `---`, write normal Markdown instructions. Prefer starting the body with `# <Readable Title>`.

- Put executable helpers only in `scripts/`.

- Put long-form documentation, schemas, API notes, and detailed examples only in `references/`.

- Put copyable templates or non-context assets only in `assets/`.

- Do not create empty placeholder files or directories.

- Do not create README, CHANGELOG, INSTALLATION_GUIDE, QUICK_REFERENCE, or any process notes.

- If you add files under `scripts/` or `references/`, `SKILL.md` must explicitly tell a future agent when to run or read them.

- Keep references one hop away from `SKILL.md`; avoid deep navigation or nested indirection.

- If an existing skill already covers the capability, update that skill instead of creating a parallel duplicate.

Design principles:

- Be concise. Add only information that is non-obvious, reusable, and worth the context cost.

- Generalize from the trace, but do not merely restate it. Infer the reusable workflow, decision points, validation steps, tool patterns, anti-patterns, and troubleshooting notes that should change a future agent's behavior.

- Prefer verifier evidence, failed tests, and concrete execution results over the agent's self-report when they conflict.

- Prefer minimal edits to the existing library over broad rewrites.

- Use progressive disclosure. Keep `SKILL.md` focused and short; move detailed material into `references/` or `scripts/` only when it improves reuse.

- Avoid duplication across `SKILL.md` and `references/`.

- Use `scripts/` for deterministic, fragile, or repeatedly rediscovered code patterns.

- Use `references/` for detailed schemas, API notes, long examples, or variant-specific details.

- Use `assets/` only for files that the agent should copy or use directly in outputs.

- The `description` field must explain both what the skill does and when to use it, including trigger contexts.

- The `SKILL.md` body should be imperative, operational, and easy to scan. Prefer workflows, decision rules, and concise examples over long prose.

- Keep `SKILL.md` under 500 lines when possible.

When interpreting the trace:

- Identify where the agent's reasoning or chosen strategy was wrong, incomplete, or too brittle.

- Treat repeated failures, failed tests, verifier mismatches, and dead-end tool choices as signals for what future agents should avoid.

- If the agent succeeded after trial and error, capture the final working pattern, the discarded bad paths, and the key decision rule that separates them.

- If the agent failed, capture the missing knowledge, validation steps, troubleshooting workflow, and the most plausible alternative approach or escalation path that should be tried earlier next time.

- Prefer decision rules such as "if X pattern appears, do Y instead of Z" over vague advice.

- If the trace does not justify a meaningful reusable change, return an empty patch and explain why.

Output requirements:

- Return exactly one JSON object with keys: `summary`, `upsert_files`, `delete_paths`.

- `upsert_files` must map relative file paths to full file contents.

- `delete_paths` must only include paths that should truly be removed as obsolete.

- Do not wrap the JSON in commentary.
\end{lstlisting}
\end{prompttemplatebox}

\begin{prompttemplatebox}[breakable]{USER\_PROMPT\_TEMPLATE}
\begin{lstlisting}[style=prompttemplate]
# Shared skill evolution task
You are updating the shared skill library rooted at the current shared-skills directory.
All file paths in the patch must be relative to that root.

## Required skill layout
When creating a new skill, use this default structure unless an existing skill for the same capability already exists:

skill-name/
|-- SKILL.md
|-- scripts/      # optional executable helpers
|-- references/   # optional reference docs loaded when needed
`-- assets/       # optional templates or files used in final outputs

`SKILL.md` must begin exactly like this, with nothing before the first `---`:

---
name: skill-name
description: Explain what the skill does and when to use it. Include trigger scenarios, file types, or task patterns.
---

# Readable Title

Only `name` and `description` are allowed in the frontmatter.

## Existing skill library
### Tree
{tree_json}

### Existing files
{files_block}

## Trial summary
- Task name: {task_name}
- Task source: {task_source}
- Verifier passed: {verifier_passed}
- Reward: {reward}
- Exception: {exception_info}
- Failed tests: {failed_tests}

## Final agent message
{final_message}

## Compacted execution trace
{trajectory_json}

## Internal workflow to follow
1. Derive the reusable capability from this trace.
2. Compare the agent's apparent plan against verifier outcomes, failed tests, and concrete tool results. Explicitly identify any wrong assumptions, brittle choices, or dead-end strategies.
3. Identify the strongest trigger phrases or task types this skill should support.
4. Extract the minimal reusable workflow, validation steps, failure-prevention guidance, and anti-patterns to avoid.
5. If the current approach failed or was brittle, infer the next-best direction, fallback, or escalation path that a future agent should try earlier, even if that exact fix was not fully executed in the trace.
6. Convert those lessons into reusable decision rules, such as when to switch tools, inspect lower-level formats, add verification earlier, or abandon a high-level API.
7. Decide whether the knowledge belongs in an existing `SKILL.md`, a new skill directory, `references/`, `scripts/`, or `assets/`.
8. Keep the patch small, high-signal, and generalized.

## Skill authoring checklist
- Do not encode task-specific filled values, IDs, or one-off outputs unless they belong in a reusable template.
- Prefer one skill directory per capability.
- If creating a new skill, create `skill-name/SKILL.md` instead of writing a bare `SKILL.md` at the library root.
- Put executable code in `skill-name/scripts/...`.
- Put long documentation or variant-specific details in `skill-name/references/...`.
- Put templates or output assets in `skill-name/assets/...`.
- If you add `scripts/` or `references/`, update `SKILL.md` so a future agent knows when to use them.
- In `SKILL.md`, put "when to use" guidance in `description`, not as a separate metadata field.
- In the body, provide:
  - a short overview of the workflow
  - clear sequential steps or a decision tree when useful
  - validation or verification steps when reusable
  - concise troubleshooting notes for the important failure modes seen in the trace
  - explicit anti-patterns or "do not rely on this when..." guidance when the trace shows a tempting but wrong path
  - a fallback or alternative direction when the trace suggests a better next attempt for future runs
  - explicit pointers to `scripts/` or `references/` if you add them
- Use imperative style.
- Keep `SKILL.md` concise. Move long or variant-specific content into `references/`.
- Only add scripts when deterministic code is genuinely reusable and brittle enough to deserve bundling.
- If the current library already contains a relevant skill, update it instead of creating a parallel duplicate.

## Output contract
Return exactly one JSON object matching this shape:

{
  "summary": "why this patch helps future runs",
  "upsert_files": {
    "skill-name/SKILL.md": "---\nname: skill-name\ndescription: what it does and when to use it\n---\n\n# Readable Title\n...",
    "skill-name/scripts/example.py": "#!/usr/bin/env python3\n...",
    "skill-name/references/details.md": "# Details\n..."
  },
  "delete_paths": []
}

Return no prose before or after the JSON object.
\end{lstlisting}
\end{prompttemplatebox}

\subsection{Skill Library Growth and Composition}

\begin{figure}[t]
    \centering
    \includegraphics[width=0.86\textwidth]{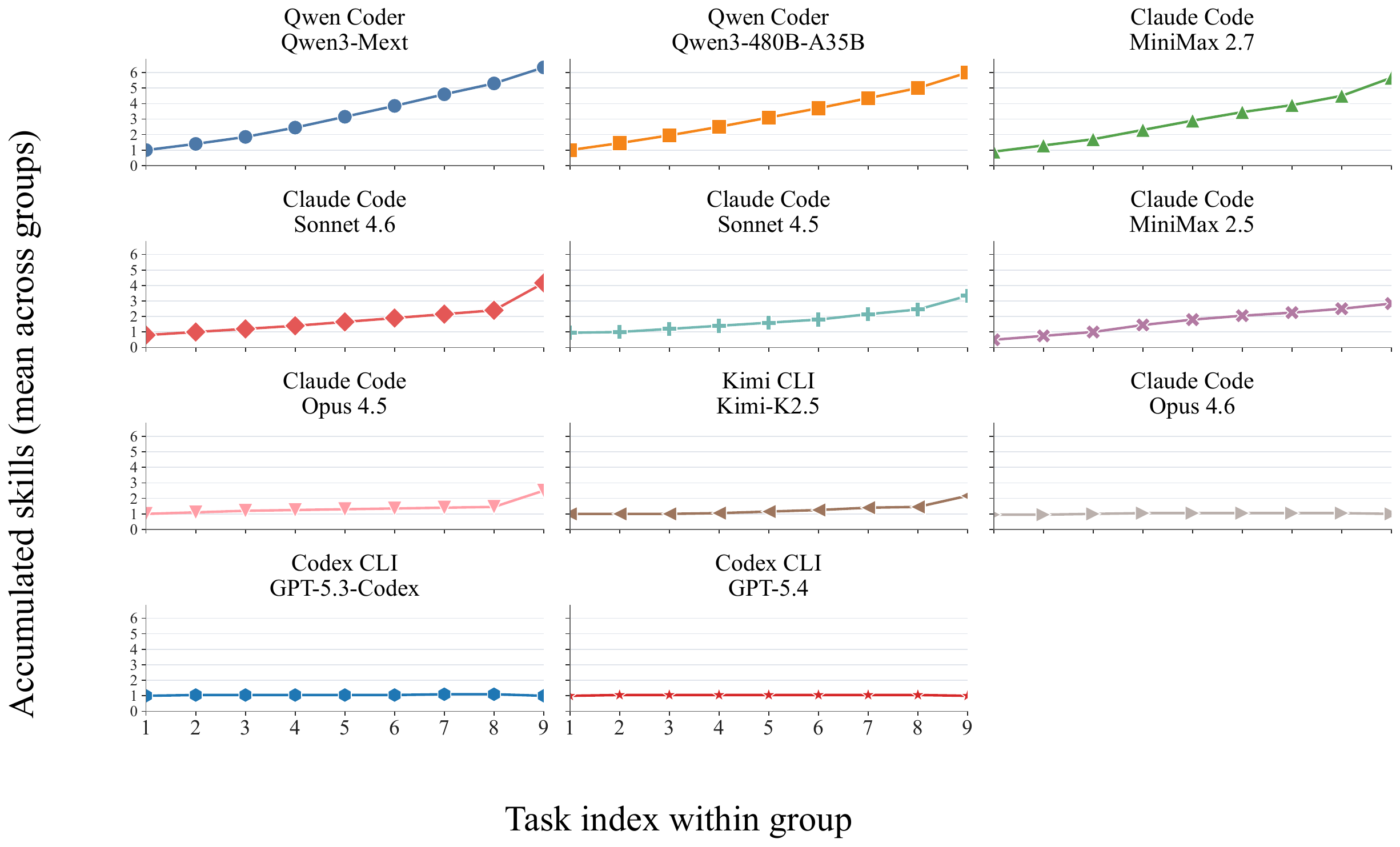}
    \captionof{figure}{Skill Count Growth by Task. The figure shows how the cumulative number of stored skills changes as the agent proceeds through sequential tasks, which helps reveal whether different models consolidate experience into a compact library or continue to expand the library over time.}
    \label{fig:skill-count-growth}

    \vspace{0.8em}

    \includegraphics[width=0.86\textwidth]{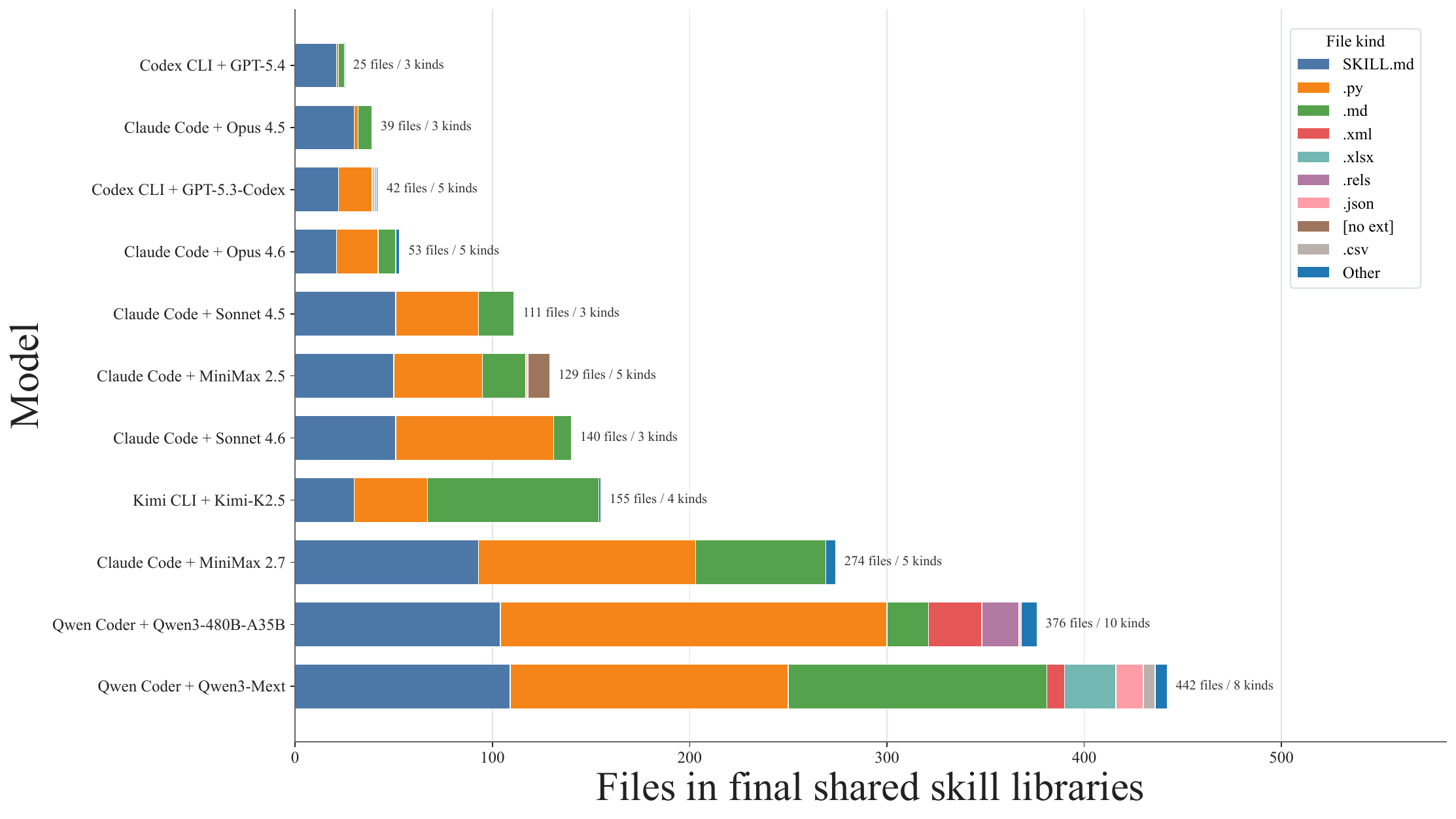}
    \captionof{figure}{Skill File-Kind Composition. The figure summarizes the composition of file kinds contained in the evolved skill libraries, which helps characterize whether different models tend to store compact procedural summaries, auxiliary files, or more fragmented collections of artifacts.}
    \label{fig:skill-file-kind-composition}
\end{figure}

Figure~\ref{fig:skill-count-growth} and Figure~\ref{fig:skill-file-kind-composition} provide two complementary views of how the external skill library evolves during sequential evaluation. The first figure tracks the cumulative number of stored skills as tasks progress, while the second summarizes the file-type composition of the resulting skill libraries.

\section{Experiments and Analysis Details}

\subsection{Metric Notes}

The main text emphasizes aggregate task completion, efficiency statistics (turns, monetary cost, and output tokens), and statistics of skill generation and reuse because these signals are consistently available across the systems reported in Table~\ref{tab:main-exp}. In Table~\ref{tab:main-exp}, Final \#Skills denotes the cumulative number of skills generated while completing a task family, and \%use denotes the percentage of tasks in which the agent reads or calls at least one previously stored skill.

\subsection{Historical-Trajectory Context Control}

To test whether the gain of the full protocol can be explained by access to longer raw histories alone, we run a single-model control on Claude Opus 4.6 in which the agent receives the full prior interaction history as additional context, but does not externalize that experience into a skill library. Table~\ref{tab:history-context-control} compares this setting against the vanilla baseline and the full protocol. In this comparison, the history-context setting reaches 47.41\% completion, which is 15.24 points below vanilla and 23.67 points below the full protocol. While this control does not isolate every factor and is limited to one model, it suggests that simply appending raw historical trajectories is not sufficient to recover the benefit of the full protocol; the observed gain is more consistent with structured externalization into reusable skills than with longer context alone.

\begin{table}[t]
    \centering
    \small
    \caption{Single-model control on Claude Opus 4.6. The history-context setting prepends the full prior interaction history as additional context, without externalizing the experience into skills.}
    \label{tab:history-context-control}
    \begin{tabular}{lcc}
        \toprule
        Setting & Completion (\%) & $\Delta$ vs. Vanilla \\
        \midrule
        Vanilla & 62.65 & -- \\
        Full historical trajectory context & 51.04 & -11.61 \\
        Full protocol (skill library) & 71.08 & +8.43 \\
        \bottomrule
    \end{tabular}
\end{table}

\subsection{Additional Efficiency Pareto Views}

\begin{figure}[t]
    \centering
    \includegraphics[width=0.82\textwidth]{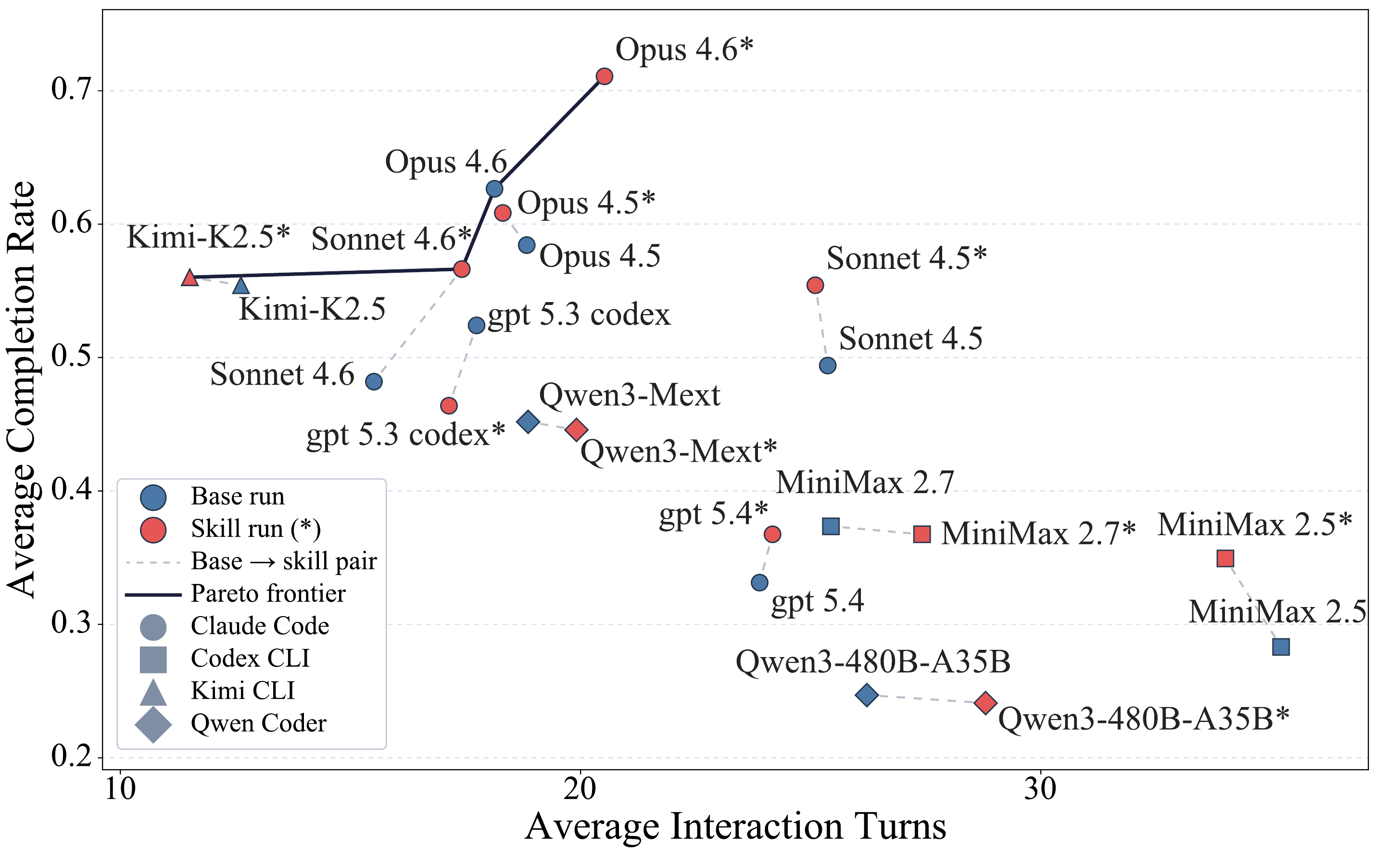}
    \caption{Completion--Interaction Turns Pareto Frontier. This view complements the main-text cost frontier by showing whether gains from skill evolution come with longer or shorter interaction sequences under the same evaluation setting.}
    \label{fig:pareto-turns}
\end{figure}

\begin{figure}[t]
    \centering
    \includegraphics[width=0.82\textwidth]{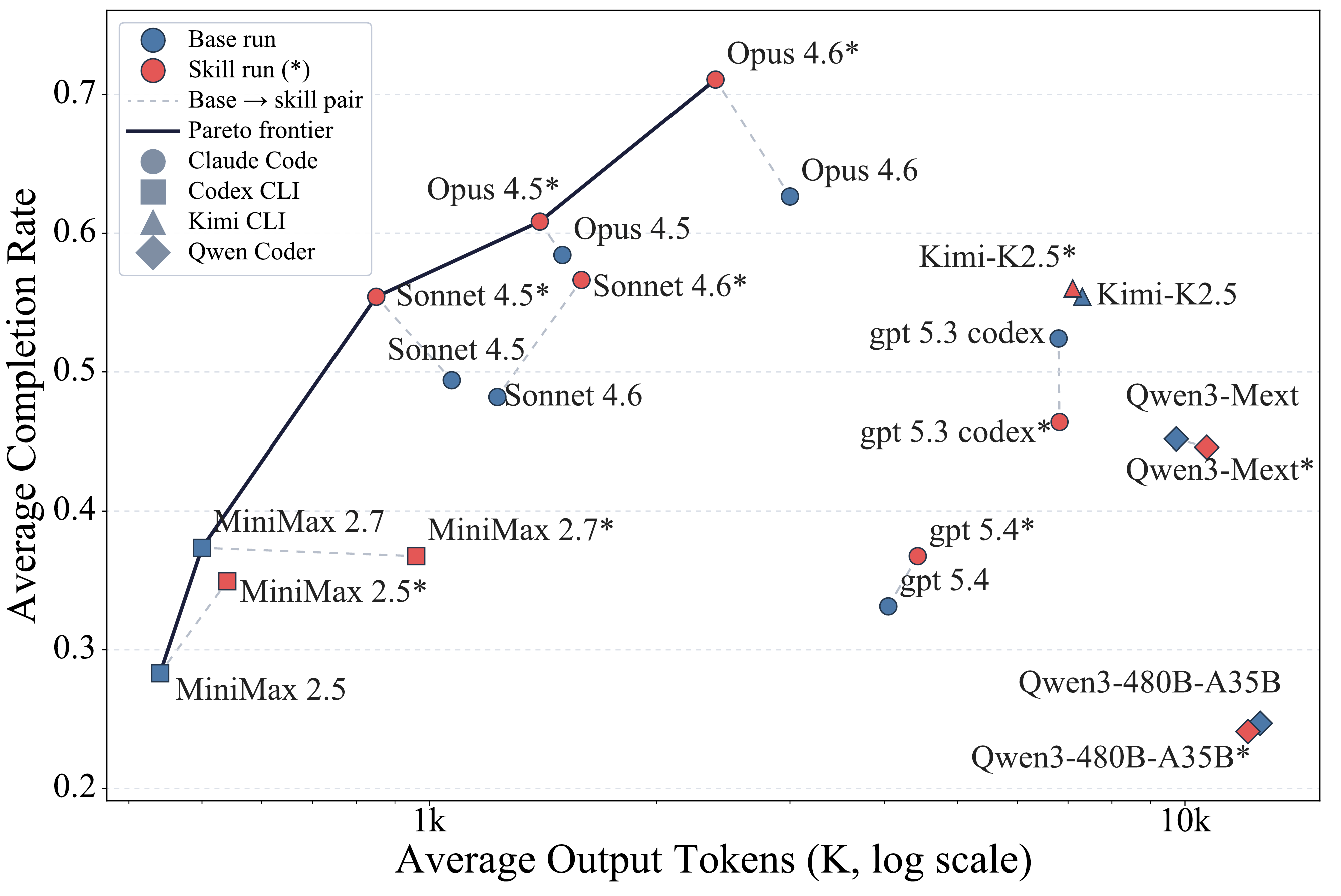}
    \caption{Completion--Output Tokens Pareto Frontier. This figure highlights the efficiency trade-off between task success and generated output volume, which is useful for diagnosing verbosity-driven gains or regressions across wrappers.}
    \label{fig:pareto-tokens}
\end{figure}

Figure~\ref{fig:pareto-turns} and Figure~\ref{fig:pareto-tokens} provide complementary Pareto views of task completion against interaction turns and output tokens.

\subsection{Full Skill-Gain Heatmap}

\begin{figure}[t]
    \centering
    \includegraphics[width=0.98\textwidth]{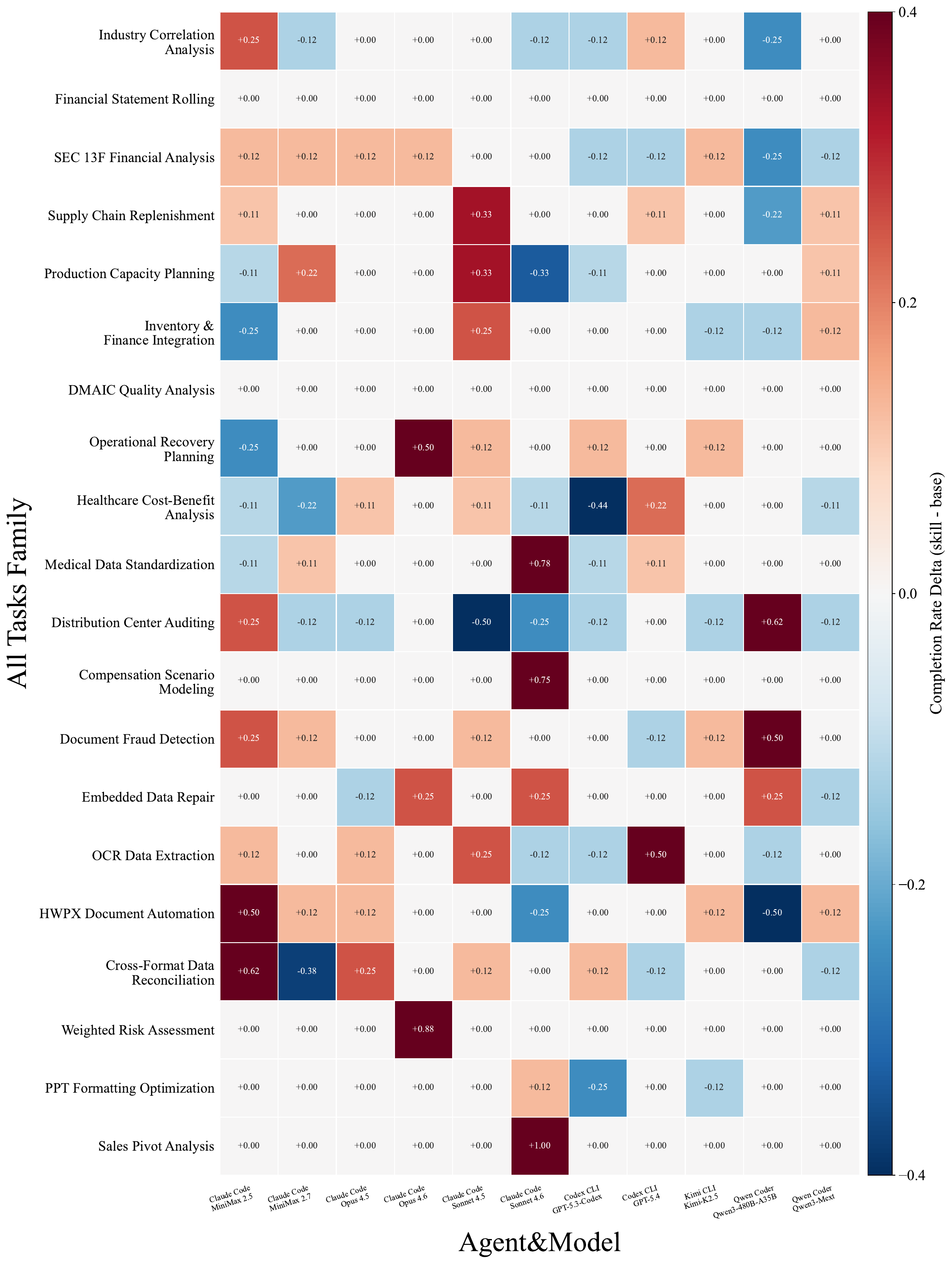}
    \caption{Family-Level Skill-Gain Heatmap. Compared with the coarse domain-grouped heatmap in the main text, this appendix figure resolves individual workflow families so that localized gains, flat regions, and regressions can be inspected in detail.}
    \label{fig:skill-gain-heatmap-all}
\end{figure}

Figure~\ref{fig:skill-gain-heatmap-all} provides a family-level complement to the coarse domain-grouped heatmap shown in the main text. It resolves the benchmark into individual workflow families and gives a more granular view of where skill evolution yields large gains, negligible changes, or regressions.

\clearpage
\section{Full Trajectory}

\makeatletter
\@ifundefined{definecolor}{\RequirePackage{xcolor}}{}
\@ifundefined{lstdefinestyle}{\RequirePackage{listings}}{}
\@ifundefined{tcbset}{\RequirePackage[most]{tcolorbox}}{}
\makeatother

\providecolor{lightgray}{gray}{0.95}
\providecolor{deepblue}{RGB}{70,130,180}
\providecolor{deepgray}{RGB}{119,136,153}

\lstdefinestyle{prompt}{
    basicstyle=\ttfamily\fontsize{7pt}{8pt}\selectfont,
    frame=none,
    breaklines=true,
    backgroundcolor=\color{white},
    breakatwhitespace=true,
    breakindent=0pt,
    escapeinside={(*@}{@*)},
    numbers=none,
    numbersep=5pt,
    xleftmargin=5pt,
    aboveskip=2pt,
    belowskip=2pt,
}

\tcbset{
  aibox/.style={
    top=10pt,
    colback=white,
    enhanced,
    center,
  }
}

\newtcolorbox{AIbox}[2][]{aibox, title=#2,#1}
\newpage
\subsection{Skill Evolution Patch}
\label{appendix:skill_patch}

\begin{AIbox}[breakable]{Patch on task:weighted-hospital-bedflow-calc}
{\color{black}\bf summary:}
\begin{lstlisting}[style=prompt]
Updates excel-formula-tasks skill with a robust XML-level cached value patching script and workflow guidance. The trace revealed that: (1) the formulas library doesn't reliably cache values in openpyxl-readable format, (2) LibreOffice is often unavailable, (3) XML-level patching is the reliable fallback but must be done carefully in a single pass to avoid corrupting formula cells, and (4) patching must handle both numeric and string formula results. The new script automates the compute-and-patch workflow that the agent discovered through trial and error.
\end{lstlisting}

{\color{deepblue}\bf upsert\_files["excel-formula-tasks/SKILL.md"]:}
\begin{lstlisting}[style=prompt]
---
name: excel-formula-tasks
description: Populate Excel workbooks with formulas using openpyxl. Use when a task requires writing INDEX/MATCH lookups, derived calculations, summary statistics, or weighted aggregations into .xlsx files. Triggered by tasks involving highlighted/yellow cells to fill, multi-sheet workbooks with source data, and formula-only requirements.
---

# Excel Formula Tasks with openpyxl

## Critical First Step: Read the Test File

Before writing any formulas, **always read the test/verifier file** (e.g., `test_output.py`) to understand:
- Whether the verifier checks formula strings or computed values
- Expected cell addresses and value tolerances
- Whether it uses `data_only=True` (expects cached values) or checks formula text
- Any structural requirements (sheet names, cell formats, etc.)

```python
import os
for root, dirs, files in os.walk('/root/'):
    for f in files:
        if 'test' in f.lower():
            print(os.path.join(root, f))
```

## Formula vs. Value Problem

**This is the #1 failure mode.** openpyxl writes formula strings but does NOT evaluate them. If the verifier loads with `data_only=True`, all formula cells return `None`.

### Decision Rule
- If verifier uses `data_only=True` or checks numeric values $\rightarrow$ you **must** also write cached values
- If verifier checks formula strings $\rightarrow$ write formulas normally
- If no test file exists $\rightarrow$ **always write cached values** as a safety measure

### Cached Value Strategy (Recommended Order)

1. **XML-level patching** (most reliable, no external deps): Write formulas with openpyxl, then use `scripts/patch_cached_values.py` to compute values in Python and inject them into the XLSX XML. **Do this in a single pass** --- multiple XML patches can corrupt formula cells.
2. **LibreOffice recalc**: If available, use `scripts/libreoffice_recalc.sh`. Check with `which libreoffice`.
3. **`formulas` library**: `pip install formulas --break-system-packages`. Note: often fails to produce openpyxl-readable cached values. **Do not rely on this as primary approach.**

### XML Patching Workflow

```bash
# After saving workbook with openpyxl:
python3 scripts/patch_cached_values.py /root/output/result.xlsx
```

See `scripts/patch_cached_values.py` for the implementation. Key points:
- Compute each formula's expected value in Python
- Patch the `<v>` element inside each `<c>` cell in the worksheet XML
- Handle both numeric values (`t` attribute absent or `"n"`) and string values (`t="str"`)
- Do all patching in **one pass** to avoid corrupting the ZIP structure
- Patch pre-existing formulas too (e.g., `=B12` reference formulas)

## Workflow

1. **Read test file** to understand verification method
2. **Examine workbook structure** --- both sheets, all rows/columns, identify:
- Yellow/highlighted cells (targets) --- check fill colors
   - Header rows with labels/years
   - Source data ranges
   - Series code columns for lookups
   - Pre-existing formulas (e.g., `=B12` in label cells)
3. **Map the layout precisely** --- off-by-one row/column errors are common
4. **Write formulas** with correct absolute/relative references:
- `$D12` --- absolute column, relative row
- `H$10` --- relative column, absolute row
- `$H$21:$L$38` --- fully absolute range
5. **Save with openpyxl** (preserves formatting)
6. **Patch cached values** using XML-level approach (single pass)
7. **Verify output** --- run `scripts/verify_formulas.py` to confirm all formula cells have cached values
8. **Run the test suite** before declaring done:
   ```bash
   cd /root && python -m pytest test_output.py -v 2>&1 | head -80
   ```

## Common Formula Patterns

### INDEX/MATCH Lookup
```python
formula = f'=INDEX(Data!$H$21:$L$38,MATCH($D{row},Data!$D$21:$D$38,0),MATCH({col}$10,Data!$H$4:$L$4,0))'
```

### Net Gap / Derived Percentage
```python
formula = f'=({col}{success_row}-{col}{fail_row})/{col}{capacity_row}*100'
```

### Summary Statistics Block
```python
stat_formulas = {
    'MIN': f'=MIN({col}{start}:{col}{end})',
    'MAX': f'=MAX({col}{start}:{col}{end})',
    'MEDIAN': f'=MEDIAN({col}{start}:{col}{end})',
    'AVERAGE': f'=AVERAGE({col}{start}:{col}{end})',
    'P25': f'=PERCENTILE({col}{start}:{col}{end},0.25)',
    'P75': f'=PERCENTILE({col}{start}:{col}{end},0.75)',
}
```

### Weighted Mean with SUMPRODUCT
```python
formula = f'=SUMPRODUCT({col}{val_start}:{col}{val_end},{col}{wt_start}:{col}{wt_end})/SUM({col}{wt_start}:{col}{wt_end})'
```

## Anti-Patterns

- **Never skip reading the test file.** The verifier determines whether you need formulas, values, or both.
- **Never assume formulas will be auto-evaluated.** openpyxl does not compute; the verifier may expect numeric results.
- **Never declare success without running `pytest`.** Self-verification with Python math is necessary but not sufficient.
- **Don't hardcode data values in formulas.** Use cell references so the workbook remains dynamic.
- **Don't modify cells outside the target range.** Preserve all existing formatting, values, and structure.
- **Don't patch XML multiple times.** Each patch reopens and rewrites the ZIP; multiple passes can corrupt formula cells by converting them to plain values. Do all patching in one pass.
- **Don't rely on `formulas` library for caching.** It often doesn't produce values readable by `data_only=True`.
- **When no test file exists, still cache values.** The verifier likely uses `data_only=True`.

## Preserving Formatting

```python
# openpyxl preserves formatting when you only set .value
wb = openpyxl.load_workbook('input.xlsx')  # NOT data_only=True for writing
ws = wb['Task']
ws['H12'] = '=INDEX(...)'  # Preserves existing fill, font, borders
wb.save('output.xlsx')
```

## Troubleshooting

| Symptom | Likely Cause | Fix |
|---------|-------------|-----|
| All formula cells show None in test | Verifier uses `data_only=True` | Patch cached values via XML |
| #REF! errors | Wrong range boundaries | Double-check row/column numbers in Data sheet |
| #N/A from MATCH | Series code mismatch | Print actual D-column values and compare |
| Wrong values | Off-by-one in row mapping | Map rows carefully to source |
| Test expects different stat order | MIN/MAX/MEDIAN row assignment wrong | Read row labels from the workbook, don't assume |
| Formulas disappeared after XML patch | Multiple XML patches corrupted cells | Redo from openpyxl save, patch once |
| `formulas` library doesn't cache | Known limitation | Use XML patching instead |
| LibreOffice not available | Not installed in container | Use XML patching instead |
| `KeyError: sharedStrings.xml` | File doesn't exist in this XLSX | List ZIP contents first, handle missing files |
\end{lstlisting}

{\color{deepblue}\bf upsert\_files["excel-formula-tasks/scripts/patch\_cached\_values.py"]:}
\begin{lstlisting}[style=prompt]
#!/usr/bin/env python3
"""Patch cached values into Excel formula cells at the XML level.

Usage: python patch_cached_values.py <xlsx_path> [--data-sheet DATA_SHEET]

This script:
1. Reads the workbook to find all formula cells
2. Computes expected values using Python (simulating INDEX/MATCH, arithmetic, stats)
3. Patches the <v> elements in the worksheet XML so data_only=True returns values

IMPORTANT: Run this ONCE after saving with openpyxl. Multiple runs may corrupt formulas.
"""
import sys
import re
import zipfile
import xml.etree.ElementTree as ET
import shutil
import os
import openpyxl
import statistics

NS = 'http://schemas.openxmlformats.org/spreadsheetml/2006/main'


def col_letter_to_index(letter):
    """Convert column letter(s) to 0-based index."""
    result = 0
    for c in letter.upper():
        result = result * 26 + (ord(c) - ord('A') + 1)
    return result - 1


def parse_cell_ref(ref):
    """Parse 'H12' into (col_letter, row_number)."""
    m = re.match(r'([A-Z]+)(\d+)', ref)
    if m:
        return m.group(1), int(m.group(2))
    return None, None


def get_cached_value(ws_data, formula, ws_task, cell_ref):
    """Attempt to compute the expected value of a formula.
    
    This handles common patterns:
    - Simple cell references: =B12, =C12
    - INDEX/MATCH lookups
    - Arithmetic: =(A1-B1)/C1*100
    - Statistics: MIN, MAX, MEDIAN, AVERAGE, PERCENTILE
    - SUMPRODUCT/SUM
    
    Returns (value, is_string) tuple.
    For complex or unrecognized formulas, returns (None, False).
    """
    if not formula or not formula.startswith('='):
        return None, False
    
    f = formula[1:]  # strip leading =
    
    # Simple same-sheet cell reference: =B12
    simple_ref = re.match(r'^([A-Z]+)(\d+)$', f)
    if simple_ref:
        col, row = simple_ref.group(1), int(simple_ref.group(2))
        val = ws_task[f'{col}{row}'].value
        is_str = isinstance(val, str) and not val.startswith('=')
        return val, is_str
    
    return None, False


def patch_xlsx_cached_values(xlsx_path, computed_values):
    """Patch cached values into XLSX XML.
    
    computed_values: dict of {cell_ref: (value, is_string)}
    where is_string indicates if the value should be stored as type="str"
    """
    tmp_path = xlsx_path + '.tmp'
    
    with zipfile.ZipFile(xlsx_path, 'r') as zin:
        # Read sheet1 XML
        sheet_xml = zin.read('xl/worksheets/sheet1.xml')
        
        ET.register_namespace('', NS)
        # Preserve all other namespaces
        namespaces = {}
        for event, elem in ET.iterparse(xlsx_path.replace('.xlsx', '_ns_dummy.xml') if False else 
                                         __import__('io').BytesIO(sheet_xml), events=['start-ns']):
            prefix, uri = elem
            if prefix:
                ET.register_namespace(prefix, uri)
                namespaces[prefix] = uri
        
        tree = ET.parse(__import__('io').BytesIO(sheet_xml))
        root = tree.getroot()
        
        for row_elem in root.iter(f'{{{NS}}}row'):
            for cell_elem in row_elem.iter(f'{{{NS}}}c'):
                ref = cell_elem.get('r')
                if ref in computed_values:
                    value, is_string = computed_values[ref]
                    if value is None:
                        continue
                    
                    # Find or create <v> element
                    v_elem = cell_elem.find(f'{{{NS}}}v')
                    if v_elem is None:
                        # Insert <v> after <f> if present
                        f_elem = cell_elem.find(f'{{{NS}}}f')
                        v_elem = ET.SubElement(cell_elem, f'{{{NS}}}v')
                    
                    if is_string:
                        cell_elem.set('t', 'str')
                        v_elem.text = str(value)
                    else:
                        # Remove string type if present
                        if cell_elem.get('t') == 's':
                            cell_elem.set('t', 'n')
                        v_elem.text = str(value)
        
        # Write modified XLSX
        modified_xml = ET.tostring(root, xml_declaration=True, encoding='UTF-8')
        
        with zipfile.ZipFile(tmp_path, 'w', zipfile.ZIP_DEFLATED) as zout:
            for item in zin.infolist():
                if item.filename == 'xl/worksheets/sheet1.xml':
                    zout.writestr(item, modified_xml)
                else:
                    zout.writestr(item, zin.read(item.filename))
    
    shutil.move(tmp_path, xlsx_path)


def main():
    if len(sys.argv) < 2:
        print(f"Usage: {sys.argv[0]} <xlsx_path>")
        sys.exit(1)
    
    xlsx_path = sys.argv[1]
    
    # Load workbook to read formulas and compute values
    wb = openpyxl.load_workbook(xlsx_path)
    ws = wb.active or wb.worksheets[0]
    
    computed = {}
    for row in ws.iter_rows():
        for cell in row:
            if isinstance(cell.value, str) and cell.value.startswith('='):
                val, is_str = get_cached_value(None, cell.value, ws, cell.coordinate)
                if val is not None:
                    computed[cell.coordinate] = (val, is_str)
    
    if computed:
        patch_xlsx_cached_values(xlsx_path, computed)
        print(f"Patched {len(computed)} cached values")
    else:
        print("No simple formulas to patch. Use task-specific computation logic.")
        print("See SKILL.md for the XML patching pattern to adapt.")


if __name__ == '__main__':
    main()
\end{lstlisting}

{\color{deepblue}\bf upsert\_files["excel-formula-tasks/references/xml-caching-pattern.md"]:}
\begin{lstlisting}[style=prompt]
# XML-Level Cached Value Patching Pattern

When openpyxl writes formulas, the resulting XLSX has `<c>` elements with `<f>` (formula) but no `<v>` (cached value). Verifiers using `data_only=True` see `None`.

## The Pattern

```python
import zipfile
import xml.etree.ElementTree as ET
import shutil
import io

NS = 'http://schemas.openxmlformats.org/spreadsheetml/2006/main'

def patch_cached_values(xlsx_path, cell_values):
    """
    cell_values: dict mapping cell ref (e.g., 'H12') to (value, is_string)
    """
    tmp_path = xlsx_path + '.tmp'
    
    with zipfile.ZipFile(xlsx_path, 'r') as zin:
        sheet_xml = zin.read('xl/worksheets/sheet1.xml')
        ET.register_namespace('', NS)
        
        # Register all namespaces from the XML
        for event, elem in ET.iterparse(io.BytesIO(sheet_xml), events=['start-ns']):
            prefix, uri = elem
            if prefix:
                ET.register_namespace(prefix, uri)
        
        tree = ET.parse(io.BytesIO(sheet_xml))
        root = tree.getroot()
        
        for row_elem in root.iter(f'{{{NS}}}row'):
            for cell_elem in row_elem.iter(f'{{{NS}}}c'):
                ref = cell_elem.get('r')
                if ref in cell_values:
                    value, is_string = cell_values[ref]
                    if value is None:
                        continue
                    
                    v_elem = cell_elem.find(f'{{{NS}}}v')
                    if v_elem is None:
                        v_elem = ET.SubElement(cell_elem, f'{{{NS}}}v')
                    
                    if is_string:
                        cell_elem.set('t', 'str')
                        v_elem.text = str(value)
                    else:
                        if cell_elem.get('t') in ('s',):
                            cell_elem.set('t', 'n')
                        v_elem.text = str(value)
        
        modified_xml = ET.tostring(root, xml_declaration=True, encoding='UTF-8')
        
        with zipfile.ZipFile(tmp_path, 'w', zipfile.ZIP_DEFLATED) as zout:
            for item in zin.infolist():
                if item.filename == 'xl/worksheets/sheet1.xml':
                    zout.writestr(item, modified_xml)
                else:
                    zout.writestr(item, zin.read(item.filename))
    
    shutil.move(tmp_path, xlsx_path)
```

## Critical Rules

1. **Single pass only.** Each patch reopens the ZIP. Multiple patches can corrupt formula cells by losing `<f>` elements.
2. **Compute ALL values in Python first**, then patch all at once.
3. **Handle string vs numeric types.** String formula results (e.g., `=B12` referencing text) need `t="str"`. Numeric results should not have `t="s"`.
4. **List ZIP contents first** if you need `sharedStrings.xml` --- it may not exist.
5. **Sheet filename:** Usually `xl/worksheets/sheet1.xml` for the first sheet. Check the workbook's `xl/_rels/workbook.xml.rels` if unsure.

## Computing Formula Values in Python

For INDEX/MATCH lookups, build a lookup dict from the Data sheet:

```python
# Build lookup: (series_code, year) -> value
lookup = {}
for row in range(data_start, data_end + 1):
    code = ws_data[f'D{row}'].value
    for col in ['H', 'I', 'J', 'K', 'L']:
        year = ws_data[f'{col}{header_row}'].value
        lookup[(code, year)] = ws_data[f'{col}{row}'].value
```

For statistics, use Python's `statistics` module or numpy:

```python
import statistics
vals = [computed[f'{col}{r}'] for r in range(start, end+1)]
results = {
    'MIN': min(vals),
    'MAX': max(vals),
    'MEDIAN': statistics.median(vals),
    'AVERAGE': statistics.mean(vals),
}
# PERCENTILE requires numpy or manual interpolation
import numpy as np
results['P25'] = float(np.percentile(vals, 25))
results['P75'] = float(np.percentile(vals, 75))
```

For SUMPRODUCT:
```python
result = sum(a*b for a,b in zip(values, weights)) / sum(weights)
```
\end{lstlisting}

{\color{deepgray}\bf delete\_paths:}
\begin{lstlisting}[style=prompt]
[]
\end{lstlisting}
\end{AIbox}

\newpage

\subsection{Weighted Hospital Bedflow Trajectory}
\label{appendix:success_traj}

This appendix presents the weighted hospital bedflow interaction trace reformatted using the same boxed dialogue style as the appendix example. The task is an Excel formula completion problem. The trajectory is notable because the agent first follows a reasonable spreadsheet-editing workflow, then encounters the well-known \emph{formula-vs-cached-value} issue, diagnoses that intermediate XML patching corrupted formula cells, and finally restarts from the original workbook to produce a verified result.

\subsubsection*{Session Metadata}
\begin{systembox}
\small
\begin{tabular}{@{}ll@{}}
\textbf{Schema:} & ATIF-v1.2 \\
\textbf{Session ID:} & 74ec9714-0797-40f3-a551-7cc95d295127 \\
\textbf{Agent:} & \texttt{claude-code} \\
\textbf{Version:} & 2.1.80 \\
\textbf{Model:} & \texttt{claude-opus-4.6} \\
\textbf{Working Directory:} & \path{/root} \\
\textbf{Git Branch:} & HEAD \\
\end{tabular}
\end{systembox}

\subsubsection*{Task Instruction}
\begin{instructionbox}
\small
You need to update \path{/root/data/workbook.xlsx}. The workbook already contains the sheets \texttt{Task} and \texttt{Data}; work only inside these existing sheets.

\textbf{Step 1:} Populate the yellow cells in \texttt{H12:L17}, \texttt{H19:L24}, and \texttt{H26:L31} on sheet \texttt{Task} with spreadsheet formulas. Use two inputs for every lookup: the series code in column \texttt{D} of the current row and the year in row \texttt{10}. The source records are on sheet \texttt{Data} in rows \texttt{21:38}. Use one of these lookup patterns: \texttt{VLOOKUP} with \texttt{MATCH}, \texttt{HLOOKUP} with \texttt{MATCH}, \texttt{XLOOKUP} with \texttt{MATCH}, or \texttt{INDEX} with \texttt{MATCH}.

\textbf{Step 2:} In \texttt{H35:L40}, calculate \texttt{Net patient flow} for the six hospitals with:
\[
(\text{Patient Admissions} - \text{Patient Discharges}) / \text{Effective Bed Capacity} * 100
\]
Then calculate the column-wise minimum, maximum, median, simple mean, 25th percentile, and 75th percentile in \texttt{H42:L47}.

\textbf{Step 3:} In \texttt{H50:L50}, calculate the weighted mean for \texttt{Metro Hospital Network (MHN)} with \texttt{SUMPRODUCT}, using the Step 2 percentages as values and the \texttt{Effective Bed Capacity} block in \texttt{H26:L31} as weights.

Keep the existing workbook formatting unchanged. Do not add sheets, macros, VBA, external links, or helper tabs. Save the finished workbook to \path{/root/output/result.xlsx}.
\end{instructionbox}

\subsubsection*{Conversation Trajectory}

% 1
\begin{userbox}
\textbf{[1]} You need to update \path{/root/data/workbook.xlsx}. The workbook already contains the sheets \texttt{Task} and \texttt{Data}; work only inside these existing sheets.

Step 1: Populate the yellow cells in \texttt{H12:L17}, \texttt{H19:L24}, and \texttt{H26:L31} on sheet \texttt{Task} with spreadsheet formulas. Use two inputs for every lookup: the series code in column \texttt{D} of the current row and the year in row \texttt{10}. The source records are on sheet \texttt{Data} in rows \texttt{21:38}. Use one of these lookup patterns: \texttt{VLOOKUP} with \texttt{MATCH}, \texttt{HLOOKUP} with \texttt{MATCH}, \texttt{XLOOKUP} with \texttt{MATCH}, or \texttt{INDEX} with \texttt{MATCH}.

Step 2: In \texttt{H35:L40}, calculate \texttt{Net patient flow} for the six hospitals with this formula:
\texttt{(Patient Admissions - Patient Discharges) / Effective Bed Capacity * 100}

Then calculate the column-wise minimum, maximum, median, simple mean, 25th percentile, and 75th percentile in \texttt{H42:L47}.

Step 3: In \texttt{H50:L50}, calculate the weighted mean for \texttt{Metro Hospital Network (MHN)} with \texttt{SUMPRODUCT}, using the Step 2 percentages as values and the \texttt{Effective Bed Capacity} block in \texttt{H26:L31} as weights.

Keep the existing workbook formatting unchanged. Do not add sheets, macros, VBA, external links, or helper tabs. Save the finished workbook to \path{/root/output/result.xlsx}.
\end{userbox}

% 2
\begin{assistantbox}
\textbf{[2]} \emph{Reasoning:} Let me start by reading the Excel workbook to understand its structure before writing any formulas.
\end{assistantbox}

% 3
\begin{userbox}
\textbf{[3]} \textit{Skill reminder injected by environment.} Base directory for this skill: \path{/etc/claude-code/.claude/skills/excel-formula-tasks}.

The reminder emphasizes two things: read the verifier/test file first, and be careful that \texttt{openpyxl} writes formulas but does not evaluate them, so a verifier using \texttt{data\_only=True} may see \texttt{None} unless cached values are present.
\end{userbox}

% 4
\begin{assistantbox}
\textbf{[4]} \textit{Tool Calls (1):} \\
\texttt{Skill("excel-formula-tasks", args=<full task instruction>)}
\end{assistantbox}

% 5
\begin{toolbox}
\textbf{[5]} \textit{Tool Response:} \\
\texttt{Launching skill: excel-formula-tasks} \\
\texttt{[metadata] \{"success": true, "commandName": "excel-formula-tasks"\}}
\end{toolbox}

% 6
\begin{assistantbox}
\textbf{[6]} \emph{Reasoning:} Let me start by examining the workbook structure and any test files to understand what's expected.
\end{assistantbox}

% 7
\begin{assistantbox}
\textbf{[7]} \textit{Tool Calls (1):} \\
\texttt{TodoWrite(} \\
\texttt{  Read test file and examine workbook structure [in progress],} \\
\texttt{  Write lookup formulas in H12:L17, H19:L24, H26:L31 [pending],} \\
\texttt{  Write Net patient flow formulas in H35:L40 and stats in H42:L47 [pending],} \\
\texttt{  Write weighted mean formulas in H50:L50 [pending],} \\
\texttt{  Run tests and verify output [pending]} \\
\texttt{)}
\end{assistantbox}

% 8
\begin{toolbox}
\textbf{[8]} \textit{Tool Response:} \\
\texttt{Todos have been modified successfully.}
\end{toolbox}

% 9
\begin{assistantbox}
\textbf{[9]} \textit{Tool Calls (2):} \\
\texttt{Bash: ls -la /root/ /root/data/ /root/output/} \\
\texttt{Bash: find /root -name '*.py' -o -name '*.xlsx' | head -30}
\end{assistantbox}

% 10
\begin{toolbox}
\textbf{[10]} \textit{Tool Response Summary:} \\
The agent confirms the presence of:
\begin{itemize}[leftmargin=*]
    \item the input workbook at \path{/root/data/workbook.xlsx},
    \item an empty output directory \path{/root/output/},
    \item and a verifier script at \path{/root/.claude/skills/excel-formula-tasks/scripts/verify_formulas.py}.
\end{itemize}
\end{toolbox}

% 11
\begin{assistantbox}
\textbf{[11]} \textit{Tool Calls (1):} \\
\texttt{Read(/root/.claude/skills/excel-formula-tasks/scripts/verify\_formulas.py)}
\end{assistantbox}

% 12
\begin{toolbox}
\textbf{[12]} \textit{Tool Response Summary:} \\
The verifier loads the workbook twice:
\begin{itemize}[leftmargin=*]
    \item once normally to inspect formulas,
    \item once with \texttt{data\_only=True} to inspect cached values.
\end{itemize}
It warns whenever a formula cell has no cached value. This confirms that the agent must preserve both formulas and cached values in the final workbook.
\end{toolbox}

% 13
\begin{assistantbox}
\textbf{[13]} Now let me examine the workbook structure in detail.
\end{assistantbox}

% 14
\begin{assistantbox}
\textbf{[14]} \textit{Tool Calls (2):} \\
\texttt{Bash: inspect Task sheet} \\
\texttt{Bash: inspect Data sheet}
\end{assistantbox}

% 15
\begin{toolbox}
\textbf{[15]} \textit{Tool Response Summary:} \\
On sheet \texttt{Task}, the agent identifies:
\begin{itemize}[leftmargin=*]
    \item lookup targets in \texttt{H12:L17}, \texttt{H19:L24}, and \texttt{H26:L31},
    \item derived metric cells in \texttt{H35:L40},
    \item summary statistic cells in \texttt{H42:L47},
    \item weighted mean cells in \texttt{H50:L50},
    \item and year headers in \texttt{H10:L10} corresponding to 2020--2024.
\end{itemize}

On sheet \texttt{Data}, rows \texttt{21:38} contain the source records, while row \texttt{4} contains the year headers.
\end{toolbox}

% 16
\begin{assistantbox}
\textbf{[16]} Now I have a clear picture. Let me also check if there's a test file.
\end{assistantbox}

% 17
\begin{assistantbox}
\textbf{[17]} \textit{Tool Calls (1):} \\
\texttt{Bash: find /root -name 'test\_output*' -o -name 'test\_*.py'}
\end{assistantbox}

% 18
\begin{toolbox}
\textbf{[18]} \textit{Tool Response Summary:} \\
No task-specific \texttt{test\_output.py} file is found. The agent therefore proceeds using the verifier script plus direct workbook inspection.
\end{toolbox}

% 19
\begin{assistantbox}
\textbf{[19]} \textit{Tool Calls (1):} \\
\texttt{Bash: write formulas into workbook and save /root/output/result.xlsx}
\end{assistantbox}

% 20
\begin{toolbox}
\textbf{[20]} \textit{Tool Response Summary:} \\
The agent writes formulas as follows:
\begin{itemize}[leftmargin=*]
    \item \textbf{Step 1 lookup:}
\begin{verbatim}
=INDEX(Data!$H$21:$L$38,
       MATCH($Drow,Data!$D$21:$D$38,0),
       MATCH(H$10,Data!$H$4:$L$4,0))
\end{verbatim}
    \item \textbf{Step 2 net patient flow:}
\begin{verbatim}
=(H12-H19)/H26*100
\end{verbatim}
    \item \textbf{Step 2 statistics:} \texttt{MIN}, \texttt{MAX}, \texttt{MEDIAN}, \texttt{AVERAGE}, and \texttt{PERCENTILE(...,0.25/0.75)}
    \item \textbf{Step 3 weighted mean:}
\begin{verbatim}
=SUMPRODUCT(H35:H40,H26:H31)/SUM(H26:H31)
\end{verbatim}
\end{itemize}
The workbook is saved to \path{/root/output/result.xlsx}.
\end{toolbox}

% 21
\begin{assistantbox}
\textbf{[21]} \textit{Tool Calls (1):} \\
\texttt{TodoWrite(} \\
\texttt{  Read test file and examine workbook structure [completed],} \\
\texttt{  Write all formulas and save workbook [completed],} \\
\texttt{  Run verification and validate output [in progress]} \\
\texttt{)}
\end{assistantbox}

% 22
\begin{toolbox}
\textbf{[22]} \textit{Tool Response:} \\
\texttt{Todos have been modified successfully.}
\end{toolbox}

% 23
\begin{assistantbox}
\textbf{[23]} Now let me verify the output thoroughly.
\end{assistantbox}

% 24
\begin{assistantbox}
\textbf{[24]} \textit{Tool Calls (1):} \\
\texttt{Bash: open /root/output/result.xlsx and inspect formulas/values}
\end{assistantbox}

% 25
\begin{toolbox}
\textbf{[25]} \textit{Tool Response Summary:} \\
The agent checks formula placement and sees that formulas exist in the expected regions. It then continues with deeper validation of cached values and workbook internals.
\end{toolbox}

% 26
\begin{assistantbox}
\textbf{[26]} \textit{Tool Calls (several):} \\
\texttt{Bash: inspect XLSX archive internals} \\
\texttt{Bash: list XML files inside result.xlsx} \\
\texttt{Bash: patch workbook XML to preserve cached values} \\
\texttt{Bash: re-open workbook and inspect selected cells}
\end{assistantbox}

% 27
\begin{toolbox}
\textbf{[27]} \textit{Tool Response Summary:} \\
During this stage, the agent discovers that the workbook has inline strings and no \texttt{sharedStrings.xml}. It attempts XML-level patching to keep cached values while preserving formulas, but the intermediate result is fragile.
\end{toolbox}

% 28
\begin{assistantbox}
\textbf{[28]} No shared strings file. The strings are inline. Let me fix the \texttt{B35:C40} cells with a simpler approach.
\end{assistantbox}

% 29
\begin{assistantbox}
\textbf{[29]} \textit{Tool Calls (1):} \\
\texttt{Bash: patch affected cells and save workbook again}
\end{assistantbox}

% 30
\begin{toolbox}
\textbf{[30]} \textit{Tool Response Summary:} \\
The agent repairs display cells such as \texttt{B35:C40} and continues checking whether formulas and cached values both survive the patching procedure.
\end{toolbox}

% 31
\begin{assistantbox}
\textbf{[31]} \textit{Tool Calls (1):} \\
\texttt{Bash: perform more detailed verification on Task sheet formula cells}
\end{assistantbox}

% 32
\begin{toolbox}
\textbf{[32]} \textit{Tool Response Summary:} \\
The verification reveals a problem: some Step 2 and Step 3 cells are no longer stored as formulas after XML patching. The agent recognizes that the patching process has partially corrupted the intended structure.
\end{toolbox}

% 33
\begin{assistantbox}
\textbf{[33]} It seems the XML patching converted the Step 2/3 formula cells to just cached values (the formulas got dropped). The issue is that the XML patching process may have corrupted the formula cells. Let me redo the entire process more carefully.
\end{assistantbox}

% 34
\begin{assistantbox}
\textbf{[34]} \textit{Tool Calls (1):} \\
\texttt{Bash: restart from original workbook and rebuild everything from scratch}
\end{assistantbox}

% 35
\begin{toolbox}
\textbf{[35]} \textit{Tool Response Summary:} \\
The agent reloads \path{/root/data/workbook.xlsx} and reconstructs the full solution pipeline:
\begin{itemize}[leftmargin=*]
    \item build a lookup dictionary from \texttt{Data!D21:L38},
    \item rewrite all Step 1 formulas with \texttt{INDEX+MATCH},
    \item rewrite all Step 2 formulas for net patient flow,
    \item rewrite all Step 2 statistics formulas,
    \item rewrite all Step 3 weighted mean formulas,
    \item then save a fresh \path{/root/output/result.xlsx}.
\end{itemize}
\end{toolbox}

% 36
\begin{assistantbox}
\textbf{[36]} \textit{Rebuilt Step 1 formula template:}
\begin{verbatim}
=INDEX(Data!$H$21:$L$38,
MATCH($D12,Data!$D$21:$D$38,0),
MATCH(H$10,Data!$H$4:$L$4,0))
\end{verbatim}

\textit{Rebuilt Step 2 formula template:}
\begin{verbatim}
=(H12-H19)/H26*100
\end{verbatim}

\textit{Rebuilt Step 3 formula template:}
\begin{verbatim}
=SUMPRODUCT(H35:H40,H26:H31)/SUM(H26:H31)
\end{verbatim}
\end{assistantbox}

% 37
\begin{assistantbox}
\textbf{[37]} \textit{Tool Calls (1):} \\
\texttt{Bash: save fresh workbook and run comprehensive verification}
\end{assistantbox}

% 38
\begin{toolbox}
\textbf{[38]} \textit{Tool Response Summary:} \\
The workbook is saved again, and the agent runs the dedicated verifier script:
\begin{verbatim}
python3 /root/.claude/skills/excel-formula-tasks/scripts/verify_formulas.py \
/root/output/result.xlsx
\end{verbatim}
\end{toolbox}

% 39
\begin{assistantbox}
\textbf{[39]} \textit{Tool Calls (1):} \\
\texttt{Bash: python3 verify\_formulas.py /root/output/result.xlsx}
\end{assistantbox}

% 40
\begin{toolbox}
\textbf{[40]} \textit{Tool Response:}
\begin{verbatim}
Sheet 'Task': 167 formula cells
  All formula cells have cached values.
Sheet 'Data': 0 formula cells
  All formula cells have cached values.
\end{verbatim}
\end{toolbox}

% 41
\begin{assistantbox}
\textbf{[41]} \textit{Tool Calls (1):} \\
\texttt{TodoWrite(} \\
\texttt{  Read test file and examine workbook structure [completed],} \\
\texttt{  Write all formulas and save workbook [completed],} \\
\texttt{  Run verification and validate output [completed]} \\
\texttt{)}
\end{assistantbox}

% 42
\begin{toolbox}
\textbf{[42]} \textit{Tool Response:} \\
\texttt{Todos have been modified successfully.}
\end{toolbox}

% 43
\begin{assistantbox}
\textbf{[43]} Final state: all required formula regions were populated, the workbook was saved to \path{/root/output/result.xlsx}, and the verifier confirmed that every formula cell in \texttt{Task} has a cached value.
\end{assistantbox}

\subsection*{Key Takeaways}
\begin{systembox}
\small
This trajectory highlights three agent behaviors that are particularly relevant for spreadsheet tasks:
\begin{enumerate}[leftmargin=*]
    \item The agent correctly inferred that workbook inspection and verifier inspection must happen before editing.
    \item The agent identified the spreadsheet-specific failure mode that formulas can exist without cached values, which breaks downstream checking under \texttt{data\_only=True}.
    \item The agent encountered a self-induced corruption caused by XML patching, explicitly diagnosed it, restarted from the original workbook, and produced a clean final artifact.
\end{enumerate}
\end{systembox}

\subsection{FAILURE TAXONOMY EXAMPLES}

\subsubsection{Verifier Toolchain Mismatch}
\begin{tcolorbox}[colback=white,colframe=black]
\noindent\textbf{Failure Mode:} Verifier / toolchain incompatibility
\\\textbf{Task:} \texttt{api-sla-at-risk-calc}
\\\textbf{Other Failure Modes:} Missing recalculation and No or Incomplete verification.
\\\textbf{Where it went wrong:}
\begin{lstlisting}[basicstyle=\tiny\ttfamily,breaklines=true]
[Verifier]
subprocess.CalledProcessError: Command
['ssconvert', '/root/output/result.xlsx',
 '/tmp/.../recalculated.xlsx', '--recalculate']
returned non-zero exit status 1.
stderr = "Unknown option --recalculate"

[Agent]
Workbook saved successfully.
Row 50: =SUMPRODUCT(H35:H40,H26:H31)/SUM(H26:H31)
\end{lstlisting}
\end{tcolorbox}

\subsubsection{Missing Cached Values}
\begin{tcolorbox}[colback=white,colframe=black]
\noindent\textbf{Failure Mode:} Missing recalculation / missing cached formula values
\\\textbf{Task:} \texttt{weighted-hospital-bedflow-calc}
\\\textbf{Other Failure Modes:} No or Incomplete verification and Premature success declaration.
\\\textbf{Where it went wrong:}
\begin{lstlisting}[basicstyle=\tiny\ttfamily,breaklines=true]
[Agent]
Step 1: Lookup formulas written
Step 2a: Net patient flow formulas written
Step 2b: Statistics formulas written
Step 3: Weighted mean formulas written
Workbook saved to /root/output/result.xlsx

[Verifier]
H12: expected 612.5, got None
...
H35: expected 4.499999999999997, got None
\end{lstlisting}
\end{tcolorbox}

\subsubsection{Incomplete Verification}
\begin{tcolorbox}[colback=white,colframe=black]
\noindent\textbf{Failure Mode:} No or Incomplete verification
\\\textbf{Task:} \texttt{weighted-port-throughput-calc}
\\\textbf{Other Failure Modes:} Verifier / toolchain incompatibility and Missing recalculation.
\\\textbf{Where it went wrong:}
\begin{lstlisting}[basicstyle=\tiny\ttfamily,breaklines=true]
[Agent]
H35 (ALP 2019): (302.5 - 395.4) / 502.0 * 100 = -18.5060
H36 (BET 2019): (111.9 - 169.1) / 203.6 * 100 = -28.0943
H38 (DEL 2019): (413.2 - 284.6) / 568.7 * 100 = 22.6130
All formulas reference correct rows and will compute correctly in Excel.

[Verifier]
Unknown option --recalculate
Command '['ssconvert', '/root/output/result.xlsx', ...,
'--recalculate']' returned non-zero exit status 1.
\end{lstlisting}
\end{tcolorbox}

\end{document}